\newif\ifconfver
\confverfalse      

\newif\ifcutshort      
\cutshorttrue

\newif\ifcutshortlvltwo  
\cutshortlvltwofalse

\ifconfver
\documentclass[10pt,twocolumn,twoside]{IEEEtran}
\else
\documentclass[11pt, draftclsnofoot, onecolumn]{IEEEtran}
\fi

\usepackage[utf8]{inputenc}
\usepackage{titlesec}
\usepackage{amsmath,graphicx}
\usepackage{arydshln}
\usepackage{graphicx}
\usepackage{psfrag}
\usepackage{subfigure}
\usepackage{url}
\usepackage{stfloats}
\usepackage{amsfonts,amssymb,amsmath,amsbsy,bm,theorem,ifthen,color}
\usepackage{array}
\usepackage{arydshln}
\usepackage{calc}
\usepackage{subfig}
\usepackage{diagbox}
\usepackage{longtable}
\usepackage{multirow}
\usepackage{multicol}
\usepackage{algorithmic,algorithm}
\usepackage{graphics,booktabs,epsfig,enumerate}
\usepackage{afterpage}
\usepackage{float}
\usepackage{cite}
\usepackage{xcolor}
\usepackage{todonotes}
\usepackage{lipsum}
\usepackage{verbatim}
\usepackage{mathtools}
\usepackage{colortbl}

{\begin{list}{}{
\settowidth{\labelwidth}{\mbox{\textnormal{#1}}}%
\setlength{\leftmargin}{\labelwidth+\labelsep}}}%
{\end{list}}

\newcommand\Nc{\ensuremath{\mathcal{N}}}

\newcommand\Dc{\ensuremath{\mathcal{D}}}

\newcommand\Sc{\ensuremath{\mathcal{S}}}

\newcommand\Bc{\ensuremath{{\mathcal{B}}}}

\newcommand\xb{\ensuremath{{\bf x}}}

\newcommand\wb{\ensuremath{{\bf w}}}
\newcommand\yb{\ensuremath{{\bf y}}}

\newcommand\ab{\ensuremath{{\bm a}}}

\newcommand\bb{\ensuremath{{\bm b}}}

\newcommand\cb{\ensuremath{{\bf c}}}

\newcommand\gb{\ensuremath{{\bm g}}}

\newcommand\ob{\ensuremath{{\bm o}}}

\newcommand\lambdab{\ensuremath{{\bm \lambda}}}

\newcommand\zerob{\ensuremath{{\bm 0}}}

\newtheorem{Lemma}{Lemma}

\newtheorem{Theorem}{Theorem}
\newtheorem{Def}{Definition}

\newtheorem{Ass}{Assumption}
\theorembodyfont{\rmfamily}

\newtheorem{Remark}{Remark}
\definecolor{orange}{RGB}{255,107,0}

\allowdisplaybreaks[4]

\DeclareMathOperator{\topk}{top}
\DeclareMathOperator{\randk}{rand}

\begin{document}

\title{Privacy-preserving Federated Primal-Dual Learning for Non-convex and Non-smooth Problems with Model Sparsification}

\author{Yiwei~Li,  Chien-Wei~Huang, Shuai~Wang,   Chong-Yung~Chi,~\IEEEmembership{Life Fellow,~IEEE,}
Tony Q. S.~Quek,~\IEEEmembership{Fellow,~IEEE}
\thanks{This paper was partially presented  at  IEEE International Workshop on
Machine Learning for Signal Processing (MLSP), Rome, Italy, Sep. 17–20,  2023~\cite{li2023privacyPDM}. This work is supported by the Ministry of Science and Technology, Taiwan,
under   Grants MOST 111-2221-E-007-035-MY2 and MOST 110-2221-E-007-031.  It is also supported in part by the National Research Foundation, Singapore and Infocomm Media Development Authority under its Future Communications Research \& Development Programme.     }
\IEEEcompsocitemizethanks{\IEEEcompsocthanksitem Y. ~Li, C.-W.~Huang  and C.-Y.~Chi are with Institute of Communications Engineering, National Tsing Hua University, Hsinchu 30013, Taiwan (e-mail: lywei0306@foxmail.com,~ s110064501@gmail.com,~cychi@ee.nthu.edu.tw).\protect
\IEEEcompsocthanksitem S. Wang is with the National Key Laboratory on Communications,
University of Electronic Science and Technology of China, Chengdu,
611731, China (e-mail: shuaiwang@link.cuhk.edu.cn).
\protect
\IEEEcompsocthanksitem T. Q. S.~Quek is  with the Singapore University of Technology and Design, Singapore 487372, and also with the Yonsei Frontier Lab, Yonsei University, South Korea (e-mail: tonyquek@sutd.edu.sg).  }}

\maketitle
\begin{abstract}	
Federated learning (FL) has been recognized as a rapidly growing research area, where the model is trained over massively distributed clients under the orchestration of a parameter server (PS)  without sharing clients' data. This paper delves into a class of federated problems characterized by non-convex and non-smooth loss functions, that are prevalent in FL applications but challenging to handle due to  their intricate non-convexity and non-smoothness nature and the conflicting requirements on communication efficiency and privacy protection.
In this paper, we propose a novel federated primal-dual algorithm with bidirectional model sparsification tailored for non-convex and non-smooth FL problems, and  differential privacy is applied for privacy guarantee. Its unique insightful properties and some privacy and convergence analyses are also presented as the FL algorithm design guidelines. Extensive experiments on real-world data are conducted to demonstrate the effectiveness of the proposed algorithm and much superior performance than some state-of-the-art FL algorithms, together with the validation of all the analytical results and properties.

\vspace{0.15cm}
\\\\
\end{abstract}

\section{Introduction}\label{sec:Introduction}
\IEEEPARstart{F}{ederated}  learning (FL)  is focused on the machine learning (ML) scenario,  where
the training data are distributed over many clients  without sharing their private data~\cite{mcmahan2017communication},  aiming to collaboratively train the global ML model under the orchestration of a parameter server (PS).  { FL has been proposed for a variety of Internet of Things (IoT) applications.  For instance, multiple IoT devices can act as  clients to
communicate with the PS for performing neural network training in intelligent IoT networks~\cite{lim2020federated,wu2020personalized,imteaj2021survey,nguyen2021federated}, including smart healthcare, smart transportation, Unmanned Aerial Vehicles (UAVs), wireless edge network~\cite{saha2020fogfl}, etc.
Nevertheless, FL also poses several key challenges, such as high communication cost and privacy protection. To be specific,}
the communication cost is certainly a major concern during the training process simply due to exchanges of the model parameters between clients and the PS~\cite{kairouz2021advances}. Moreover, although local data are not exposed directly, privacy protection~\cite{li2019convergence} is still needed as clients’
sensitive information can be reversely deduced by professional or experienced adversaries~\cite{li2020secure,li2022federated} through model inversion
attacks~\cite{fredrikson2015model,geiping2020inverting}, or differential attacks \cite{dwork2014algorithmic,bagdasaryan2020backdoor} during the training process. Therefore, the high communication cost and privacy protection are still major concerns so far in practical FL implementation~\cite{xia2021survey}.

{For improving the communication efficiency, many works~\cite{wang2023beyond,wang2023batch,liang2019variance}  relied on the promising strategies including:} 1) partial client participation, which reduces the communication cost per round by allowing only partial clients to participate in the training process; 2) local stochastic gradient descent (SGD), which increases the number of local inner iterations  at each round for reducing the total communication rounds; 3) transmission of compressed model, which cuts down the communication cost by   model quantization~\cite{wang2021quantized,reisizadeh2020fedpaq} or model sparsification~\cite{sattler2019robust,hu2022federated}  provided that the synchronization between the clients and the PS is achieved.
{Furthermore, communication cost can also be reduced by designing faster convergence rate algorithms. It is well-known that
the primal-dual method (PDM)  has favorable  convergence properties and widespread utilization in distributed optimization problems~\cite{hong2016convergence,hajinezhad2016nestt}.}  There has been a growing interest in designing and analyzing FL algorithms based on PDM, specially,  several recent works~\cite{li2023privacyPDM,li2022federated,karimireddy2020scaffold,zhang2021fedpd,zhou2023federated}  have successfully applied PDM to FL, demonstrating its superiority over state-of-the-art FL algorithms like FedAvg~\cite{li2019convergence,li2022federated, zhang2021fedpd} in   both communication efficiency and learning performance.  Notably, the PDM has been  acknowledged as a powerful optimization method for non-convex   problems in centralized  scenarios~\cite{hong2016convergence}, {but it is still challenging to solving the non-convex and non-smooth (NCNS) FL problems~\cite{li2022federated,ding2019stochastic}.  }

{For providing data privacy in FL, differential privacy (DP) techniques   have gained significant popularity in recent years thanks to its complete theoretic guarantees,} algorithmic simplicity and negligible system overhead~\cite{li2022network,truex2020ldp}.
However, the main bottleneck of  DP-based approaches in FL is the  tradeoff between data privacy and learning performance, often referred to as the privacy-utility tradeoff.   Existing studies~\cite{mcmahan2017learning,erlingsson2019amplification,li2022network,geyer2017differentially} have demonstrated serious performance degradation resulting from DP noise required to achieve a specific level of privacy protection. The performance degradation can be further exacerbated by factors such as repeated synchronization~\cite{dennis2021heterogeneity} and increasing model size~\cite{wang2019collecting}.  To handle this challenge, recent works in DP-based FL have focused on balancing the tradeoff between privacy protection and learning performance through noise reduction strategies~\cite{li2022network,li2021deep}, including privacy amplification~\cite{DPFedC2023,erlingsson2019amplification,balle2018privacy} and model sparsification~\cite{hu2022federated,ahn2022model}.

{Despite their improved results, to the best of our knowledge,  none of them can effectively handle the NCNS FL problems in one work,  especially under uncompromising requirements for both communication efficiency and privacy protection~\cite{Hong2017StochasticPG,hajinezhad2016nestt,hong2016convergence}.
In this work, by integrating the PDM to non-convex and non-smooth FL problem, two novel communication-efficient FL algorithms are proposed with DP applied for privacy protection. }

\subsection{Related Work}\label{sec:related}
Numerous FL works have focused on DP-based approaches to provide privacy protection on clients' sensitive data. For instance, the DP-FedAvg algorithm~\cite{li2020secure, wei2019performance,li2022federated} successfully reduces communication costs by allowing   \emph{local SGD} and \emph{partial client participation}. However, this algorithm may not be suitable for  handling  FL problems when involving non-smooth regularizers. Other works, such as \cite{zhang2021fedpd}, have developed PDM algorithms for solving non-convex but smooth FL problems, while the works~\cite{li2022federated,huang2019dp}  have extended PDM in order to handle  non-smooth but convex FL problems.  However, to the best of our knowledge, none of existing works can effectively handle the issues of both non-convexity and non-smoothness while also considering  privacy protection and communication-efficiency issues.

On the other hand, many existing works  have invested  tremendous efforts in enhancing  communication efficiency, by  extending upon FedAvg to further reduce either the total communication rounds or the communication cost per round.
The majority of these efforts attempted to leverage advanced optimization techniques to expedite the convergence rate of the developed DP-based FL algorithms. These works include FedProx~\cite{li2020federated}, FedNova~\cite{wang2020tackling}, SCAFFOLD~\cite{karimireddy2020scaffold}, FedDyn~\cite{durmus2021federated}, etc.
Another line to cut down the communication cost is to transmit the compressed model under synchronization with PS.  Several recent  studies~\cite{wang2021quantized,wang2023batch,reisizadeh2020fedpaq,amiri2020federated}  have delved into model quantization methods for lowering communication cost. However, these approaches inevitably induce  quantization errors and  incompatibility with DP techniques.  Meanwhile, the works~\cite{wang2018atomo,sattler2019robust, sattler2019sparse, ahn2022model,hu2022federated} have explored model sparsification by introducing two prominent sparsification techniques, namely top-$k$ and rand-$k$, that seamlessly align with  DP  techniques. In most instances, the top-$k$ method tends to outperform the rand-$k$ approach in terms of learning performance. However, the former suffers from  the  ``curse of primal averaging", where in  the aggregated global model by the PS comes up with a dense solution~\cite{yuan2021federated}.
To mitigate this issue,  the work~\cite{ahn2022model} proposed a shared pattern sparsification technique by applying the same sparsification pattern for gradient estimation across all participating clients.  However,  the majority of the aforementioned works overlooked the critical aspect of privacy preservation. While some recent studies~\cite{hu2022federated,agarwal2018cpsgd,zhang2020private,kerkouche2021compression}  have simultaneously  considered communication efficiency and
privacy protection, their primary focus has been on FedAvg or its variants. As far as we are aware,  none of the existing works have ventured upon solving non-convex and non-smooth FL problems while simultaneously considering communication efficiency and privacy protection, hence motivating us to develop advanced privacy-preserving federated PDM algorithms under the practical FL scenario.

\subsection{Contributions}
The main contributions of this paper are summarized as follows:

\begin{itemize}

\item  Two novel DP-based FL algorithms using PDM for {solving NCNS FL problem}  are proposed, including one referred to as DP-FedPDM without involving model sparsification,  and the other considering model sparsification, referred to as bidirectional sparse DP-FedPDM (BSDP-FedPDM), together with their unique properties (cf. (P1) through (P4) in Section III-B). The former, DP-FedPDM, can be regarded as  a fundamental FL algorithm, or the version of the latter with the model sparsification switched off.



\item  Comprehensive privacy and convergence analyses for DP-FedPDM are proposed, that can be used as guidelines for the FL algorithm design, including the order of the required number of communication rounds  in $\mathcal{O}(1/\zeta)$ (cf. Remark \ref{remark:rmk2}), which, to the best of our knowledge, is the lowest for achieving a $\zeta$-stationary solution to {the FL problem.}


\item  Extensive experimental results using real-world datasets have been provided to validate the effectiveness of the proposed FL algorithms, including the verification of all the analytical results and properties, as well as the demonstration of their  much superior performance over some state-of-the-art algorithms.
\end{itemize}

\vspace{0.1cm}

{\bf Synopsis:} {For ease of the ensuing presentation, all the mathematical notations used are listed in Table \ref{tab:Notations}}.  Section~\ref{sec:Preliminaries} introduces the preliminaries of FL and DP.   Section~\ref{sec:DP based FedAvg} presents the DP-FedPDM and  BSDP-FedPDM algorithms.  Section~\ref{sec:Convergence analysis} provides the privacy analysis and convergence analysis of the former algorithm.  Experimental results are presented in Section \ref{sec:Experimental Details}, and Section \ref{sec:Conclusions} concludes the paper.
\vspace{0.1cm}
\begin{table}[h]
\renewcommand{\arraystretch}{1.25}
\centering
\caption{Notations}
\begin{tabular}{m{2.1cm}|m{5.9cm}}
\hline
\hline
\textbf{Notation} &  \textbf{Description} \\
\hline
${\mathbb R}^d$, ${\mathbb R}^{m \times n}$ & Set of real $d\times 1$ vectors and  $m\times n$ matrices.\\
\cdashline{1-2}[0.8pt/1pt]
$[N]$ & Integer set $\{1,\ldots, N\}$.\\
\cdashline{1-2}[0.8pt/1pt]
$|\Dc|$ & Size  of  set $\Dc$.\\
\cdashline{1-2}[0.8pt/1pt]
$\{\xb_i \}$ & Set of $\xb_1,\xb_2,\dots$ for all the admissible $i$.\\
\cdashline{1-2}[0.8pt/1pt]
${\bf x}^{\top}$ & Transpose of vector ${\bf x}$.\\
\cdashline{1-2}[0.8pt/1pt]
${\bf I}_d$ & $d\times d$ identity matrix.\\
\cdashline{1-2}[0.8pt/1pt]
$[{\bf X}]_{jk}$ & $(j,k)$-th entry of matrix ${\bf X}$.\\
\cdashline{1-2}[0.8pt/1pt]
$\|{\bf x}\|$,   $\|{\bf x}\|_1$, $\|{\bf x}\|_0$ & Euclidean norm, $\ell_{1}$-norm and the cardinality  of vector ${\bf x}$, respectively.\\
\cdashline{1-2}[0.8pt/1pt]
$\nabla f( \xb )$ & Gradient of function $f( \xb )$ with respect to (w.r.t.) $\xb$.\\
\cdashline{1-2}[0.8pt/1pt]
$\mathbb{E}[\cdot]$, $\mathbb{P} [\cdot]$ & Statistical expectation and the probability function, respectively.\\
\hline
\hline
\end{tabular}
\label{tab:Notations}
\end{table}


\section{Preliminaries}\label{sec:Preliminaries}
\subsection{Federated Learning}\label{subsec:FedAvg}
Let us consider a vanilla FL framework that consists one PS and $N$ clients as illustrated in Fig. \ref{fig:FL_system}. Suppose that each client $i \in [N]$ holds a  private local dataset $\mathcal{D}_i$.
Then FL problem  can be expressed as
\vspace{0.1cm}
{
\begin{figure}[t]
\begin{center}
\resizebox{0.85\linewidth}{!}{\hspace{-0cm}\includegraphics{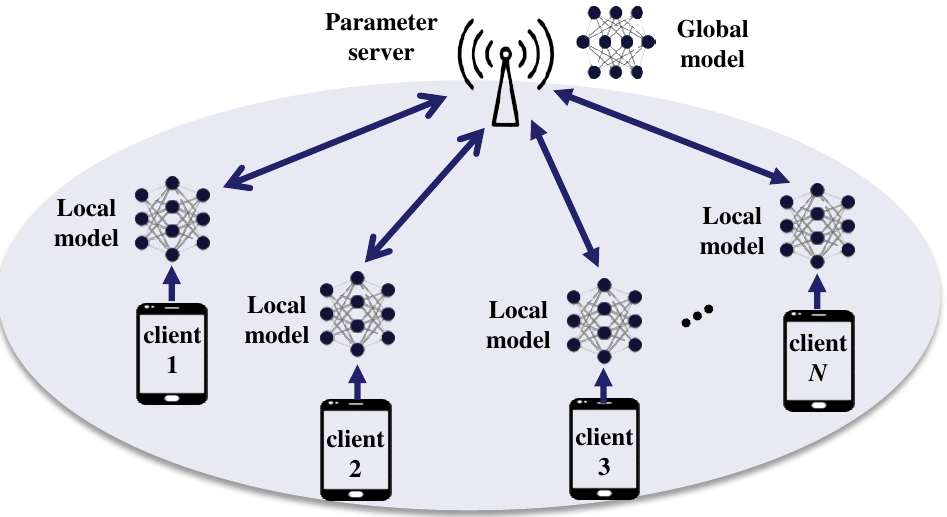}}
\end{center}
\vspace{-0.17cm}
\caption{The framework of vanilla FL system.}
\label{fig:FL_system}
\end{figure}
\vspace{-0.1cm}
\begin{subequations} \label{eqn:problem_2}
\begin{align}
\min& ~  \big\{\frac{1}{N}  \sum_{i=1}^{N}  f_{i}(\xb_{i})  + h(\xb_{0}) \big\}  \label{eqn:p3_a} \\
\text { s.t.}&~   \xb_{i}=\xb_{0}, \quad \forall i \in [N], \label{eqn:p3_b}
\end{align}
\end{subequations}}where $\xb_i \in \mathbb{R}^d$ is the model of client $i$ (local model), and $\xb_0 \in \mathbb{R}^d$ is the global model, {and $N$ is the number of devices (or clients), $f_{i}$ is a non-convex loss function,  and $ h$ is a non-smooth (possibly non-convex) regularizer including the regularization parameter.  Problem~\eqref{eqn:problem_2} is pervasive and covers many FL applications in IoT applications, e.g., sparse learning~\cite{li2022federated,chen2018federated}. However, the existing FL algorithms are almost not directly applicable
when the loss function of problem \eqref{eqn:problem_2} is non-convex and non-smooth \cite{li2019convergence,li2022federated}.} This motivates us to develop a   communication-efficient and privacy-preserving FL algorithm for solving problem \eqref{eqn:problem_2}.

\vspace{0.02cm}
\subsection{Differential Privacy}\label{subsec:Privacy Concern}
In FL system, the adversary can be ``honest-but-curious'' server or clients in the system. The adversaries may be curious about a target client’s private data and intend to steal  from the shared messages. Furthermore, some clients may collude with the PS to extract private information about a specific   client.   These attackers can eavesdrop all the shared messages during the execution of the training rather than actively inject false messages into or interrupt message transmissions.
The widely used $(\epsilon, \delta)$-DP mechanism for the privacy protection is defined as follows.

\vspace{0.05cm}
\begin{Def}\label{Def:DP defintion}  ($(\epsilon, \delta)$-DP~\cite{dwork2014algorithmic}). Suppose that $\cal X$ is a given dataset, and there exist two neighboring datasets  $\mathcal{D},\mathcal{D}^{\prime} \subset \mathcal{X}$  such that $|{\cal D}\cup{\cal D'}|-|{\cal D}\cap{\cal D'}|=1$. A randomized mechanism $\mathcal{M}: \mathcal{X} \rightarrow \mathbb{R}^d $ achieves  $(\epsilon, \delta)$-DP if for any subset of outputs ${\cal O} \subseteq  Range(\mathcal{M})$:
\begin{align}\label{def DP}
\!\!\!	\mathbb{P}[\mathcal{M}(\mathcal{D}) \in \mathcal{O}] \leq \exp(\epsilon)  \cdot \mathbb{P}\left[\mathcal{M}\left(\mathcal{D}^{\prime}\right) \in \mathcal{O} \right]+\delta,
\end{align} where $\epsilon > 0$ and $\delta \in [0,1).$
\end{Def}

Note that a smaller $\epsilon$ means stronger privacy protection, and $\delta$ stands for the probability to break the $(\epsilon, 0)$-DP. The  $(\epsilon, \delta)$-DP can be implemented by properly adding Gaussian noise vector $\boldsymbol{\xi}\in{\mathbb R}^d$ (i.e., DP noise) to protect data privacy \cite{dwork2014algorithmic}, that is
%
%
\vspace{0.05cm}
\begin{align}\label{eqn: DP gaussian}
\mathcal{M}(\Dc) = \gb(\Dc) + \boldsymbol{\xi},~ \,\boldsymbol{\xi}  \sim \Nc(\zerob,\sigma^2{\bf I}_d),
\end{align}where   $\gb$ is a specified query function, and $\Nc(\zerob,\sigma^2{\bf I}_d)$  is the distribution of a zero-mean Gaussion noise with covariance matrix  $\sigma^2 {\bf I}_d$.  The following lemma provides the required ``noise variance" $\sigma^2$ (i.e., variance of every element in $\boldsymbol{\xi}$) for the $(\epsilon, \delta)$-DP mechanism given by \eqref{eqn: DP gaussian}.
\vspace{-0.1cm}
\begin{Lemma}  \label{Lemma: global sensitivity}
\cite[Theorem 3.22]{dwork2014algorithmic}  Suppose that the randomized mechanism ${\cal M}$ satisfies $(\epsilon,\delta)$-DP defined in \eqref{eqn: DP gaussian}. Then, the minimum required noise variance $\sigma^2$ is given by
\begin{align}\label{eqn:noise_variance}
\sigma^2 = \frac{2 s^{2} \ln(1.25/\delta)}{\epsilon^{2}},
\end{align}where $ s$,  the  $\ell_2$-norm sensitivity  of  $\gb$ in \eqref{eqn: DP gaussian}, is given by
{\small \begin{align}\label{eqn:global sensitivity_f}
s \triangleq \max _{ \mathcal{D},\mathcal{D}^{\prime} \subset {\cal X}}\big\|  \gb(\mathcal{D})-  \gb\big(\mathcal{D}^{\prime}\big)\big\|,
\end{align}}in which ${\cal X}$ is the domain of function $\gb$.
\end{Lemma}
{
\begin{Def}\label{Def:Total privacy loss} (Privacy loss~\cite{dwork2014algorithmic}). Suppose that a randomized mechanism $\mathcal{M}$ satisfies $(\epsilon, \delta)$-DP. Let ${\cal D}$ and ${\cal D}^\prime$ be two neighboring datasets and   $\ob$ be a possible output of ${\mathcal{M}} (\Dc)$ and ${\mathcal{M}} ({\Dc}^{\prime})$.  The privacy loss is defined by
\begin{align} \label{eqn:def2}
c (\ob )= \ln \Big(\frac{ \mathbb{P}\big[ {\mathcal{M}} (\Dc)= \ob  \big]}{\mathbb{P}\big[\mathcal{M}( {\Dc}^{\prime})= \ob \big]}\Big).	
\end{align}
\end{Def}

Note that when $\ob$ is a continuous random vector, $\mathbb{P}[\cdot]$ stands for its probability density function, and this is exactly the case in our work.
\vspace{0.2cm}
\section{ Proposed Privacy-preserving Federated Primal-dual Method}\label{sec:DP based FedAvg}
\subsection{Proposed Primal-dual Method in FL}
The federated PDM (FedPDM) solves (\ref{eqn:problem_2}) by successively and iteratively searching for the desired model, a saddle point of the augmented Lagragian \eqref{eqn: augmented Lagrangian_1}, w.r.t. $\xb$, $\xb_0$ (minimization) and $\lambdab$ (maximization):

\vspace{-0.5cm}
\begin{small}
\begin{subequations}
\begin{align}
&\!\!\!\mathcal{L} \big(\xb, \xb_{0}, \lambdab\big) = \frac{1}{N}\sum_{i=1}^{N} \mathcal{L}_{ i}\big(\xb_{i}, \xb_{0}, \lambdab_{i} \big) + h(\xb_{0}), \label{eqn: augmented Lagrangian_1} \\
&\!\!\!\mathcal{L}_{i}\big(\xb_{i}, \xb_{0}, \lambdab_{i}\big) =  f_{i}\big(\xb_{i}\big)   +   \lambdab_{i}^\top (\xb_{0} -  \xb_{i})   +  \frac{\rho}{2}  \left\| \xb_{0}-\xb_{i} \right\|^{2}, \label{eqn: augmented Lagrangian_2}
\end{align}
\end{subequations}
\end{small}in which  $\xb =[\xb_{1}^{\top},\ldots, \xb_{N}^{\top}]^{\top}$, $\lambdab=[\lambdab_{1}^{\top},\ldots, \lambdab_{N}^{\top}]^{\top}$, $\rho > 0$ is the penalty parameter which must be large enough such that ${\cal L}(\cdot)$ is strongly convex in $\xb_{i}$~\cite{07a7d8b8158648d59974a856e79ecdf3}. Note that, $\xb_i$ (primal variable) and $\lambdab_i$ (dual variable) are updated {at the local site of client $i$, } while $\xb_{0}$ denotes the global model updated at the central PS.
\vspace{0.1cm}

Furthermore, we consider \emph{local SGD}  and   \emph{partial client participation}    strategies to improve communication efficiency~\cite{li2019convergence}. Then,
the update rule for the proposed FedPDM  is as follows:

\vspace{-0.35cm}
\begin{subequations}\label{eq: sadmm-update}
\begin{align}
\xb_{i}^{t,r+1} =& \xb_i^{t,r} -  \eta^t \big( \nabla f_i(\xb_{i}^{t,r};\mathcal{B}_i^{t,r}) - \lambdab_{i}^t +\rho(\xb_{i}^{t,r}-\xb_{0}^t) \big), \label{eq: sadmm-update_1} \\
\xb_{i}^{t+1} =&  \xb_{i}^{t,Q_i^t},   \label{eq: sadmm-update_2}\\
\lambdab_{i}^{t+1} =&  \lambdab_{i}^{t} + \rho\left(\xb_{0}^{t}-\xb_{i}^{t+1}\right),   \label{eq: sadmm-update_3}\\
{\yb}_{i}^{t+1} =&  \xb_{i}^{t+1} -  \frac{\lambdab_{i}^{t+1}}{\rho},   \label{eq: sadmm-update_y} \\
\xb_0^{t+1} =& {\rm prox}_{\rho^{-1} h}\Big( \frac{1}{K} \sum_{i \in \mathcal{S}_t} {\yb}_{i}^{t+1} \Big), \label{eq: sadmm-update_5}
\end{align}
\end{subequations}where $\eta^t$ denotes the step size; $\mathcal{B}_i^{t,r} \subseteq {\cal D}_i$ is a mini-batch dataset with $|\mathcal{B}_i^{t,r}|=b$;  $\mathcal{S}_t \subseteq  [N]$  denotes  the set of participated clients at the $t$-th round   with  $|\mathcal{S}_t| = K$;  $Q_i^t$ represents the number of local SGD iterations
(where, rather than a preassigned parameter, $Q_i^t$ is determined automatically by the algorithm under consideration);
$\operatorname{prox}_{\rho^{-1} h} ({\bf u})$ is the proximal operator~\cite{parikh2014proximal} defined by
\begin{align}\label{eqn:solution_prox}
\operatorname{prox}_{\rho^{-1} h} ({\bf u})  \triangleq  \arg \min_{{\bm x}} \{ h ({\bm x})+\frac{\rho}{2} \big\|{\bm x}- {\bf u} \big\|^{2} \}.
\end{align}
The proof of~\eqref{eq: sadmm-update_5} is given in Appendix~\ref{appdixA}.
For a preassigned small $\nu>0$ for controlling the size of the vector $(\xb_{i}^{t,r+1}-\xb_{i}^{t,r})$, the inner loop ends when
\begin{align} \label{eqn:localacc}
\big\| \nabla f_i(\xb_{i}^{t,r}; \mathcal{B}_i^{t,r})-\lambdab_{i}^t+\rho(\xb_{i}^{t,r}-\xb_{0}^t)\big\|^2\leq \nu,
\end{align}and $Q_i^t$ determined by \eqref{eqn:localacc} is the smallest number of iterations spent in the inner loop.\vspace{0.1cm}
\begin{algorithm}[!t]
\caption{\!\!{\bf :}  DP-FedPDM}
\begin{algorithmic}[1]\label{alg:ADMM2}
\STATE \textbf{Input:}  System parameters $b$,   $T$, $\nu$, $\rho$, $\eta^t$, $K.$
\STATE Initialize  $\xb_{0}^{0}$,  $\{\xb_{i}^{0,0}\}$,  $\{\lambdab_{i}^{0}\}$, and $\mathcal{S}_{0}=[N]$.
\FOR{$t=0, 1, \ldots, T-1$}
\STATE $\textbf{Client side:}$
\FOR{$i \in \mathcal{S}_{t}$ in parallel}
\STATE Set $\xb_{i}^{t,0}= \xb_{0}^{t}$.
\FOR {$r =0,1,\ldots $ }
\STATE   Sample mini-batch $\mathcal{B}_{i}^{t,r}$  from $\mathcal{D}_{i}$.
\STATE  Update $\xb_{i}^{t,r+1}$  using \eqref{eq: sadmm-update_1}.
\IF {\eqref{eqn:localacc} is  satisfied}
\STATE  Set $Q_i^t=r+1$ and go to line 13.
\ENDIF
\ENDFOR
\STATE  Update   $\xb_{i}^{t+1}$ by \eqref{eq: sadmm-update_2}.
\STATE  Update $\lambdab_{i}^{t+1}$ by  \eqref{eq: sadmm-update_3}.
\STATE Update  $ {\bf y}_i^{t+1}$ by \eqref{eq: sadmm-update_y}.
\STATE  Compute  $\widetilde{{\yb}}_{i}^{t+1} $ by \eqref{eqn:x_0_update_a} and send it to the PS.
\ENDFOR
\STATE  $\textbf{Server side:}$
\STATE    Update   $  \xb_{0}^{t+1} $ by \eqref{eqn:x_0_update_b}.
\STATE  Update the subset of clients   ${\cal S}_{t+1} \subseteq [N]$  through  randomly sampling without replacement, and then broadcast  $ \xb_{0}^{t+1} $  to all the clients.
\ENDFOR	
\end{algorithmic}
\end{algorithm}
%
To guarantee  $(\epsilon, \delta)$-DP for the local model $\yb_i^{t+1}$ to be uploaded, the   DP noise $\boldsymbol{\xi}_{i}^{t+1} \sim \mathcal{N}\big(\bm{0}, \sigma_{i,{t+1}}^{2} \mathbf{I}_{d}\big)$   is added to $ {\yb}_{i}^{t+1}$, i.e.,
\begin{align}\label{eqn:x_0_update_a}
\widetilde{{\yb}}_{i}^{t+1}  =  {\yb}_{i}^{t+1} + \boldsymbol{\xi}_i^{t+1}   =  \xb_{i}^{t+1} -  \frac{\lambdab_{i}^{t+1}}{\rho} +\boldsymbol{\xi}_i^{t+1}.
\end{align}
Finally, $\xb_0^{t+1}$ given by \eqref{eq: sadmm-update_5} can  alternatively be expressed as

\vspace{-0.3cm}
\begin{align}\label{eqn:x_0_update_b}
\xb_0^{t+1} ={\rm prox}_{\rho^{-1} h}\Big( \frac{1}{K} \sum_{i \in \mathcal{S}_t}  \widetilde{{\yb}}_{i}^{t+1} \Big).
\end{align}

It is worth noting that the proximal operator is commonly used in optimization algorithms associated with non-differentiable loss functions. For some regularizers, \eqref{eqn:solution_prox} has a closed-form solution. For instance, when $h = \|\cdot \|_1$, it leads to a sparse and closed-form solution with soft thresholding operator~\cite{parikh2014proximal}, which can be expressed as follows.
\vspace{-0.075cm}
\begin{align}
\!\big[{\rm prox}_{\rho^{-1} h}(\xb)\big]_i  \! = \! \Bigg\{\begin{array}{ll}
\!\!\!{\rm sgn}([\xb]_i)  \max{\big\{\big|[\xb]_i\big|-\frac{1}{\rho},  0\big\}}, \!\!\!\!\!& {{\rm if} \ [\xb]_i \neq 0,}\vspace{0.1cm} \\
\!\!\!0,     & {{\rm otherwise.} }
\end{array}
\end{align}where ${\rm sgn}(x) = x/|x|$ for $x \neq  0$. The proposed differentially private FedPDM  (DP-FedPDM) is implemented by Algorithm~\ref{alg:ADMM2}.





\subsection{DP-FedPDM with  Bidirectional Model Sparsification}\label{subsec:DP-FedPDM}
To further reduce the communication cost, by adding artificial
model compression operation in the proposed DP-FedPDM, we come up with a bidirectional sparse DP-FedPDM (BSDP-FedPDM). This method transmits sparsified model parameters in both the uplink (clients to the PS) and downlink (PS to clients) communications during each round, while previous works only considered uplink model compressing. Importantly, this approach can also mitigate the adverse effects of  DP  noise since the variance of DP noise  is  proportionally to the model size~\cite{abadi2016deep}.
The $\topk_k$ or $\randk_k$ sparsifiers under consideration are defined as follows:
\vspace{-0.2cm}
\begin{Def}[$\topk_k$ and $\randk_k$  Sparsifiers~\cite{hu2022federated}]\label{def:sparsifier}
Let $\yb = [y_{1}, \dots, y_{d}]^{\top} \in \mathbb{R}^d$ and $\yb_{\Downarrow} = [y_{l_1}, \dots, y_{l_d}]^{\top}, \, |y_{l_i}| \geq |y_{l_j}| \ \forall l_i \leq l_j.$
The $\topk_k$ and $\randk_k$  sparsifier are defined as follows:
\begin{subequations}\label{spar_def}
\begin{align}
\topk_k(\yb)&\triangleq \yb_{\Downarrow 1:k} = [y_{l_1}, \dots, y_{l_k}]^{\top} \in \mathbb{R}^k,
\\
\randk_k(\yb)&\triangleq [y_{i_1}, \dots, y_{i_k}]^{\top} \in \mathbb{R}^k ,
\end{align}
\end{subequations}where $\randk_k(\yb)$ is a size-$k$ vector random sample from $\yb$.
\end{Def}For ease of later use, let us define the following notations
\vspace{-0.075cm}
\begin{subequations}
\begin{align}
{\bm I}(\topk_k(\yb)) &= [l_1, \cdots, l_k],  \
{\cal I}(\topk_k(\yb)) = \{l_1, \cdots, l_k \},
\\
{\bm I}(\randk_k(\yb)) &= [i_1, \cdots, i_k], \
{\cal I}(\randk_k(\yb)) = \{i_1, \cdots, i_k \}.
\end{align}
\end{subequations}

\vspace{-0.15cm}

In practice, $\randk_k(\yb)$ may easily omit crucial elements of $\yb,$ resulting in performance degradation, especially when $k$ is small. In contrast, $\topk_k$ preserves components of the top $k$ largest magnitudes, thus maintaining higher model accuracy than $\randk_k$.
In the following, we present the proposed BSDP-FedPDM with $\topk_k$ and $\randk_k$  sparsifiers, and its pseudo-code is provided in Algorithm~\ref{alg:algorithm_2}.

For  the $\topk_k$ scheme,  client $i \in \Sc_{t}$ at first performs model sparsification on  local model $\yb_i^{t+1}$ (cf. \eqref{eq: sadmm-update_y}),   yielding a sparsified model
\begin{align}\label{eqn:y_update_compressed_0}
{\widehat{\yb}}_{i}^{t+1} = \topk_k(\yb_{i}^{t+1}) \in \mathbb{R}^{k},
\end{align}
\noindent  and the associated index vector ${\bm I}({\widehat{\yb}}_{i}^{t+1})$, implying that the uplink compression ratio
$\alpha_{_U}\triangleq k/d$, i.e., the dimension ratio after and before model sparsification on ${\bf y}_i^{t+1}$. Then, a noise vector $(\boldsymbol{\xi}_{i}^{t+1})^{\prime} \sim \mathcal{N}\big(\bm{0}, \sigma_{i,{t+1}}^{2} \mathbf{I}_{k}\big)$ is added  to ${\widehat{\yb}}_{i}^{t+1}$  for guaranteeing $(\epsilon, \delta)$-DP, that is,
\begin{align}\label{eqn:y_update_compressed}
(\widehat{{\yb}}_{i}^{t+1})^{\prime}  =  {\widehat{\yb}}_{i}^{t+1} + (\boldsymbol{\xi}_i^{t+1})^{\prime}.
\end{align}
Next, client $i$ transmits these $k$ index-value pairs  $\mathcal{A}_i^{t+1} \triangleq ({\bm I}({\widehat{\bf y}}_{i}^{t+1}), ({\widehat{\bf y}}_{i}^{t+1})^{\prime}) , \forall i \in \Sc_{t}$ to the PS. At the PS side, after receiving $\mathcal{A}_i^{t+1}$ from client $i$,  PS at first converts $(\widehat{{\yb}}_{i}^{t+1})^{\prime}$ into a $d$-dimensional ($d$-dim) vector $({\widehat{\bf y}}_{i}^{t+1})^{\prime\prime} \in \mathbb{R}^d$  by zero padding for all the missing entries (line~\ref{global_restore}). Subsequently, the PS aggregates all the models  $({\widehat{\bf y}}_{i}^{t+1})^{\prime\prime}, \forall i$  into a complete $d$-dim model, though some zero entries still exist when $|\cup_{i\in{\cal S}_t} {\cal I}(\widehat{\bf y}_i^{t+1})|<d$.
 Inspired by \cite{zheng2023fedpse},  we adopt the element-wise aggregation scheme  to mitigate the impact of the ``curse of primal averaging". Specifically, \eqref{eqn:x_0_update_b} is replaced with the following expression:
\begin{align}\label{eqn:x_0_update_compressed}
(\xb_0^{t+1})^{\prime} ={\rm prox}_{\rho^{-1} h}\Big(  \sum_{i \in \mathcal{S}_t}  ({\widehat{\bf y}}_{i}^{t+1})^{\prime\prime} \oslash \wb^{t+1} \Big),
\end{align}
where $\oslash$ denotes the element-wise division operator and
\begin{align}
\!\!\![\wb^{t+1}]_j = \Bigg\{\begin{array}{ll}
1,  ~~{{\rm if} \   [({\widehat{\bf y}}_{i}^{t+1})^{\prime\prime}]_j = 0,}\vspace{0.1cm} \\
\sum_{i \in {\cal S}_t}\sum_{x \in {\cal I}({\widehat{\bf y}}_{i}^{t+1})} \mathbb{I}_j(x),      ~{{\rm otherwise,} }\vspace{0.1cm}
\end{array}
\end{align}in which $\mathbb{I}_j(\cdot)$ is the indicator function,  defined by
\begin{align}
\mathbb{I}_j(x) = \Bigg\{\begin{array}{ll}
1, & {{\rm if} \ x=j,}\vspace{0.1cm} \\
0,     & {{\rm otherwise.} }
\end{array}
\end{align}
To further reduce the communication cost, $\topk_k$ sparsifier  is applied to the  global model $(\xb_0^{t+1})^{\prime}$ by the PS,   yielding the sparsified model
\begin{align}\label{eqn:y_update_compressed_1}
({\widehat \xb_0}^{t+1})^{\prime} = \topk_{k}((\xb_0^{t+1})^{\prime})  \in \mathbb{R}^{k^{\prime}},
\end{align}
\noindent and the associated index vector ${\bm I}(({\widehat \xb_0}^{t+1})^{\prime})$, and thus the downlink model compression ratio  $\alpha_{_D}=k^{\prime}/d$.
Then,  PS broadcasts the index-value pair $\mathcal{A}_0^{t+1} = {({\bm I}(({\widehat \xb_0}^{t+1})^{\prime}), ({\widehat{\bf x}}_{0}^{t+1})^{\prime}) }$ to all the selected clients for next round of model update. As the selected clients receive $\mathcal{A}_0^{t+1}$,  they can recover $\widehat{\bf x}_0^{t+1}$ from
$({\widehat{\bf x}}_{0}^{t+1})^{\prime}$ through zero padding missing entries. All the other steps remain the same as given in Algorithm~\ref{alg:ADMM2}.  The preceding global aggregation and model regulation procedure at the PS  also applies to $\randk_k$.

\begin{algorithm}[t]
\caption{Proposed BSDP-FedPDM}\label{alg:algorithm_2}
\begin{algorithmic}[1]
\STATE \textbf{Input:}  System parameters $b$,   $T$, $\nu$, $\rho$, $\eta^t$, $K$,    compression ratio $\alpha_{_U}$ ($\alpha_{_D}$) for the unplink (downlink).
\FOR{$t=0, 1, \ldots, T-1$}
\STATE \textbf{Client side:}
\STATE Download $\mathcal{A}_0^{t} \triangleq ({\bm I}(({\widehat \xb_0}^{t})^{\prime}), ({\widehat{\bf x}}_{0}^{t})^{\prime}) $ from the PS.
\STATE Convert $({\widehat{\bf x}}_{0}^{t})^{\prime} $  from dimension $k^{\prime} = d\times \alpha_{_D}$ into the associated $d$-dim $\widehat{\bf x}_0^t$  with $\|\widehat{\bf x}_0^t\|_0=k'$.   \label{alg1:ln_aggre_0}
\FOR{$i \in \mathcal{S}_{t}$ in parallel}
\STATE Set $\xb_{i}^{t,0}= \widehat{\xb}_{0}^{t}$.
\STATE  Obtain ${\bf y}_i^{t+1}$ by following lines 7--16 in Algorithm~\ref{alg:ADMM2}.
\STATE  Update  ${\widehat{\yb}}_{i}^{t+1}$   by \eqref{eqn:y_update_compressed_0}.
\STATE  Compute  $(\widehat{{\yb}}_{i}^{t+1})^{\prime}$ by \eqref{eqn:y_update_compressed} for $k= d\times \alpha_{_U}$. \label{alg1:ln_noise_local}
\STATE Send the $\mathcal{A}_i^{t+1} = ({\bm I}({\widehat{\yb}}_{i}^{t+1}), ({\widehat{\bf y}}_{i}^{t+1})^{\prime})  $ to the PS. \label{alg1:ln_enc}
\ENDFOR
\STATE \textbf{Server side:}
\STATE    Convert  $({\widehat{\bf y}}_{i}^{t+1})^{\prime}, \forall i \in \Sc_t$ from dimension $k= d\times \alpha_{_U}$ into the associated $d$-dim $(\widehat{\bf y}_0^{t+1})^{\prime\prime}$ with $\|({\widehat{\bf y}}_{i}^{t+1})^{\prime\prime}\|_0$$=$$k$.  \label{global_restore}
\vspace{-0.4cm}
\STATE    Obtain   $(\xb_0^{t+1})^{\prime} $ by \eqref{eqn:x_0_update_compressed}.  \label{alg1:ln_global}
\STATE    Obtain  $({\widehat \xb_0}^{t+1})^{\prime}$ by \eqref{eqn:y_update_compressed_1} with dimension $k^{\prime}= d\times \alpha_{_D}$.
\STATE  Update the subset of clients   ${\cal S}_{t+1} \subseteq [N]$  through  randomly sampling without replacement. \label{alg1:ln_aggre}
\STATE Broadcast $ \mathcal{A}_0^{t+1} = ({\bm I}(({\widehat \xb_0}^{t+1})^{\prime}), ({\widehat{\bf x}}_{0}^{t+1})^{\prime}) $ to all selected clients. \label{alg1:ln_global_broadcast}
\ENDFOR
\end{algorithmic}
\end{algorithm}
Let us conclude this section with the following properties of the proposed BSDP-FedPDM:
\begin{itemize}
    \item[(P1)] In view of model dimension reduction resulted from the sparsifier applied, the term inside the parentheses of \eqref{eqn:x_0_update_compressed} (for global model aggregation and regularization) can be thought of as an actuarial average of each non-zero element (a true model element) of  $({\widehat{\bf y}}_{i}^{t+1})^{\prime\prime}, \forall  i\in {\cal S}_t$, thereby free from the effect of ``curse of primal averaging".
\vspace{0.1cm}
\item[(P2)]  Uplink communication cost
can be reduced by 50\% when $\alpha_{_U}=1$ (i.e., without sparsification) owing to the combined model $(\widehat{{\yb}}_{i}^{t+1})^{\prime}$  perturbed by DP noise
(cf. \eqref{eqn:x_0_update_a}), instead of $(\xb_i^{t+1},\lambdab_i^{t+1})$, while the total communication cost will be reduced to $50\% \times \alpha_{_U}$ if model sparsification under the control of the system operator is applied,  i.e., $\alpha_{_U} <1$. However, as a tradeoff with communication cost reduction, the model sparsification will also induce some performance loss.
\vspace{0.1cm}
\item[(P3)] The global aggregation and model regularization is performed by the PS (cf. \eqref{eqn:x_0_update_b} and \eqref{eqn:x_0_update_compressed}). When the non-smooth $\ell_1$-norm regularizer (i.e., $h({\bf x})=\gamma \|{\bf x}\|_1$ where $\gamma$ is a non-negative parameter) is used, the trained global model ${\bf x}^\star$ (owing to the inherent soft sparisification nature of ${\ell_1}$-norm) will be a sparse solution, thus especially suited to the $\topk_k$ sparsification scheme for larger communication cost saving. Furthermore, we found that the performance of ${\bf x}^\star$ is robust against the DP noise under a proper choice of $\gamma$, which will be justified by the experimental results
in Subsections \ref{subsec:impact_of_spar} and \ref{subsec:impact_of_nonsmooth} later.
\vspace{0.1cm}
\item [(P4)] The property (P3) also applies to DP-FedPDM simply because the proposed BSDP-FedPDM and DP-FedPDM are identical as $\alpha_{_U}=\alpha_{_D}=1$.
\end{itemize}

\section{Privacy and Convergence Analysis for DP-FedPDM}\label{sec:Convergence analysis}\vspace{-0.05cm}
\subsection{Assumptions}\label{subsec: Assumtions}

\vspace{-0.cm}
\begin{Ass}[Smoothness and lower bounded]\label{Ass: Assumption1}
Each loss function $f_{i}(\cdot)$ in \eqref{eqn:problem_2} is Lipschitz smooth, i.e., $f_i$ is continuously differentiable and there exists an $L>0$ such that
\begin{align}
\| \nabla f_{i}({\bm x}) -  \nabla f_{i}({\bm y})\big\| \leq L\big\| {\bm x} - {\bm y} \big\|, \forall  i \in [N]. \label{eqn:L_smooth}
\end{align}Besides, $\mathcal{L}(\cdot)$ given in \eqref{eqn: augmented Lagrangian_1} is bounded below, i.e.,
\begin{align}
\underline{f} \triangleq \inf_{\xb,\xb_0,\lambdab} \mathcal{L}( \xb, \xb_{0}, \lambdab ) > -\infty.
\end{align}
\end{Ass}

\vspace{-0.5cm}
\begin{Ass}[Unbiased gradient and bounded variance]\label{Ass: Assumption2}  For the mini-batch dataset $\mathcal{B}_{i}^{t,r}$ with size $|\mathcal{B}_{i}^{t,r}| = b$ at the $t$-th round, the associated mini-batch gradient  satisfies
\begin{subequations}
\begin{align}
&\mathbb{E} \left[\nabla f_{i}\big(\xb_{i}^{t,r}; \mathcal{B}_{i}^{t,r}\big)  \right] =  \nabla f_{i}\big(\xb_{i}^{t,r} \big),     \\
&\mathbb{E} \Big[ \big\|\nabla f_{i}\big(\xb_{i}^{t,r}; \mathcal{B}_{i}^{t,r}\big)-\nabla f_{i}(\xb_{i}^{t,r})\big\|^{2}\Big]  \leq {\phi^{2}}, \label{eqn:A2_2}
\end{align}
\end{subequations}{for all $t, r$.}
\end{Ass}

\vspace{-0.5cm}
\begin{Ass}[Bounded gradient]\label{Ass: Assumption3}
The mini-batch gradients $\nabla f_{i} (\xb_{i }^{t, r}; \mathcal{B}_{i}^{t, r})$, $\forall t,r$, are bounded, that is,
\begin{align}
\| \nabla f_i (\xb_{i}^{t,r}; \mathcal{B}_i^{t, r})\| \leq G.
\end{align}
\end{Ass}
{
\vspace{-0.40cm}
\subsection{Privacy Analysis}
\begin{Theorem}\label{Thm:total_budget}
With the noise vector $\boldsymbol{\xi}_{i}^t \sim \mathcal{N}\big(\bm{0}, \sigma_{i,t}^{2} \mathbf{I}_{d}\big)$ added to $\yb_i^t$ (cf. \eqref{eqn:x_0_update_a}),
the minimum noise variance {$\sigma_{i,t}^2 $ (cf. \eqref{eqn:noise_variance}) for guaranteeing $(\epsilon, \delta )$-DP is given by}
\begin{align}
\sigma_{i,t}^2  =  \frac{ 2  s_{i,t}^2 \ln \big(1.25/\delta\big)}{\epsilon^2},
\end{align}where
\vspace{-0.2cm}
\begin{align} \label{eqn:sensitivity}
s_{i,t} = \Bigg\{\begin{array}{ll}
\text{{\normalsize $\frac{1-| 1- \rho \eta^t  |^{Q_i^t}}{1-| 1- \rho \eta^t |}$}}4 \eta^t G, & {{\rm if} \ \rho \eta^t\neq 2,}\vspace{0.1cm} \\
4 \eta^t Q_i^tG,     & {{\rm otherwise.} }
\end{array}
\end{align}
\end{Theorem}
\noindent \textit{Proof:} {The proof is given in Appendix \ref{appdix: proof of Lemma}}. $\hfill \blacksquare$ }\vspace{0.1cm}

By Theorem~\ref{Thm:total_budget}, one can see that $\sigma_{i,t}^{2}$  is  larger for larger $Q_i^t$, though a larger $Q_i^t$ can improve the communication efficiency in FL system~\cite{li2022federated}, so the tradeoff between privacy protection and communication efficiency is determined by $Q_i^t$.\vspace{0.1cm}



The total privacy loss of client $i$   is the sum of all the privacy losses (cf. \eqref{eqn:def2}) over $T$ communication rounds~\cite{dwork2014algorithmic}, it can be estimated by multiple methods, e.g., the moments accountant method~\cite{abadi2016deep}, which is used in proving the following theorem.

\begin{Theorem}\label{Thm:total_privacyloss}
Let $p_i= K/N $ and $q_{i}=Q_i^t b/|\mathcal{D}_i| $  be the  fraction of participating clients  and  data used by client $i$, respectively. Under the $(\epsilon, \delta)$-DP at  round $t$, the minimum total privacy loss over $T$ communication rounds is given by
\begin{align}
\bar{\epsilon}_{i}^T &= c_{0}  q_{i}^2 \epsilon \sqrt{\frac{ p_i T}{1- q_{i}} },  \forall i \in [N], \label{eqn:total_privacyloss}
\end{align}where $ c_{0}>0$ is a constant  dependent upon $\delta$.
\vspace{-0.1cm}
\end{Theorem}
\textit{Proof:} The proof basically follows that of Theorem 1 in~\cite{li2022federated} under the case of data sampling without replacement.  \hfill $\blacksquare$\vspace{0.1cm}

On can observe from \eqref{eqn:total_privacyloss} that $\bar{\epsilon}_{i}^T$
increases linearly with  $\epsilon$, implying weaker privacy protection (i.e., larger $\epsilon$) but better learning performance. This will
be justified in experimental results later.

\vspace{-0.1cm}
\subsection{Convergence Analysis}\label{subsec: Convergence analysis for DP based FedAvg}
Motivated by~\cite{hong2016convergence}, the quantity $P(\{\xb_{i}^{t}\}, \xb_{0}^{t}, \{\lambdab_{i}^{t}\})$  used as convergence performance measure is defined by
\begin{align}\label{eqn:criterion}
P(\{\xb_{i}^{t}\}, \xb_{0}^{t}, \{\lambdab_{i}^{t}\})  \triangleq & \sum_{j=1}^{N}\Big[ \big\| \nabla_{\xb_{j}} \mathcal{L}( \{\xb_{i}^{t}\}, \xb_{0}^{t}, \{\lambdab_{i}^{t}\} )\big\|^2  + \big\| \nabla_{\lambdab_{j}}\mathcal{L}( \{\xb_{i}^{t}\}, \xb_{0}^{t}, \{\lambdab_{i}^{t}\} )  \big\|^2  \Big] \notag \\
& + \big\| \nabla_{\xb_{0}}\mathcal{L}( \{\xb_{i}^{t}\}, \xb_{0}^{t}, \{\lambdab_{i}^{t}\} ) \big\|^2.
\end{align}It can be verified that if $P(\{\xb_{i}^{t}\}, \xb_{0}^{t}, \{\lambdab_{i}^{t}\}) \rightarrow 0$ as $t$ increases, then a stationary-point solution to problem \eqref{eqn:problem_2} can be obtained~\cite{zhang2021fedpd}.

\begin{Theorem} \label{Thm:Theorem_convergence}
Suppose that Assumptions \ref{Ass: Assumption1}-\ref{Ass: Assumption3} hold and $2\sqrt{5}-4 \leq L \leq\rho/4$. Then, the following inequality holds:
\begin{align}\label{eqn:Thm1}
\!\!\!\!\frac{1}{T} \sum_{t=0}^{T-1} \mathbb{E} \big[   P(\{\xb_{i}^{t}\}, \xb_{0}^{t}, \{\lambdab_{i}^{t}\})  \big]   \leq  \zeta &  \triangleq \underbrace{  \frac{ (\mathcal{L}( \{\xb_{i}^{0}\}, \xb_{0}^{0}, \{\lambdab_{i}^{0}\} ) - \underline{f}) C_0}{T}}_{A_0}  + \underbrace{ C_1 \nu }_{A_1} + \underbrace{ C_2 \phi^2 }_{A_2}  + \underbrace{ C_3 \sigma^2 }_{A_3},
\end{align}where $\nu$, $\phi^2$ have been defined in \eqref{eqn:localacc} and Assumption \ref{Ass: Assumption2}, respectively; $\sigma^2 \triangleq \max_{i,t} \sigma_{i,t}^2;$  $C_0,$ $C_1,$ $C_2,$ $C_3$ are constants depending on system parameters $\rho$ and $L$  given in  \eqref{eqn:C4_2}.
\end{Theorem}

\noindent \textit{Proof:} {The proof is given in Appendix \ref{appenx:C}. $\hfill \blacksquare$ }

Let us conclude this section with the following two remarks based on  Theorem  \ref{Thm:Theorem_convergence}.

\begin{Remark} \label{remark:remark1}
The convergence performance bound in Theorem \ref{Thm:Theorem_convergence} is dependent upon $A_0$, $A_1$, $A_2$, $A_3$. Specifically,
i) $A_0\rightarrow 0$ as $T$ increases; ii) $A_1\rightarrow 0$ as $\nu\rightarrow 0$; iii) $A_2$ can be made arbitrarily small for $b$ large enough since $\phi^2\rightarrow 0$ as $b\rightarrow |{\cal D}_i|$~\cite{zhang2021fedpd}; iv) $A_3$ can be reduced by letting $\rho \eta^t \rightarrow 1$ or increasing $\epsilon$ by Theorem \ref{Thm:total_budget}.
\end{Remark}

\begin{Remark} (\textbf{Communication complexity}). \label{remark:rmk2}
By Theorem \ref{Thm:Theorem_convergence}, a $\zeta$-stationary solution can be achieved under the following parameter setting:
\begin{subequations}
\begin{align}
T&=4(\mathcal{L}(\{\xb_{i}^{0}\}, \xb_{0}^{0}, \{\lambdab_{i}^{0}\} ) - \underline{f})C_0/\zeta, \label{rmk2_a} \\
\nu &= \zeta/(4C_1),  \phi^2=\zeta/(4C_2), \sigma^2 =\zeta/(4C_3), \label{rmk2_b}
\end{align}
\end{subequations}where $\sigma^2 \triangleq \max_{i,t} \sigma_{i,t}^2$. It can be inferred that \eqref{rmk2_b} can be achieved by Remark \ref{remark:remark1}. Finally, we would like to emphasize that the required communication round $T$ is in $\mathcal{O}(1 / \zeta)$ by \eqref{rmk2_a}, while to the best of our knowledge, most of state-of-the-art FL algorithms require $T$ in ${\cal O}(1/\zeta^2)$
for a $\zeta$-stationary solution of  non-convex problems~\cite{noble2022differentially,tran2021feddr}.
\end{Remark}

{Based on the above analyses,  the proposed  BSDP-FedPDM algorithm  is  well-suited to IoT applications thanks to its superior convergence speed,  stronger privacy protection and    lower communication cost. This is crucial, as many IoT devices face challenges in handling large ML models under limited computation and communication resources. }

\section{Experimental Results and Discussions}\label{sec:Experimental Details}
\subsection{Experimental Model}\label{subsec:Experimental setting}
We evaluate the performance of the proposed  DP-FedPDM  by considering  the commonly known non-convex and non-smooth   logistic regression problem~\cite{zhang2021fedpd} with the loss function
{\small
\begin{align}
F({\bf X}) =\frac{1}{N}  \sum_{i=1}^{N}  f_{i}({\bf X})  + \gamma \| {\bf X} \|_1,
\end{align}}where $\| {\bf X} \|_1$ $=  \sum_{j=1}^{m}  \sum_{k=1}^{n} \left|[{\bf X}]_{jk}\right|$, ${\bf  X}=[{\bm x}_1,\dots,{\bm  x}_m]^{\top} \in \mathbb{R}^{m \times n}$,  and $\gamma \ge 0$ denotes the regularization parameter, and
\begin{align}\label{eq: crossentropy}
\!\!\!\!\!f_{i}({\bf X})=  \frac{1}{|{\cal D}_{i}|} \sum_{j=1}^{|{\cal D}_{i}|}  \Big[ \ln \big(1+\exp(- {\rm Tr}({\bf X} {\bf Y}_{ij})\big) \Big]+ \beta \sum_{j,k} \frac{ [{\bf X}]_{jk}^2}{1+ [{\bf X}]_{jk}^2},
\end{align}

\vspace{-0.1cm}
\noindent in which ${\bf Y}_{ij} = \ab_{ij}\bb_{ij}^{\top}$ is sparse, since $\ab_{ij} \in \mathbb{R}^n$ represents the feature vector and $\bb_{ij}\in\mathbb{R}^m$ satisfying $\|\bb_{ij}\|_0=1$ denotes the label vector of the $j$-th sample in $\Dc_i$. The concavity of $f_i$ increases as $\beta \ge 0$ increases. Note that the regularizer $\|{\bf X}\|_1$ (convex envelope of $\|{\bf X}\|_0$~\cite{07a7d8b8158648d59974a856e79ecdf3})  is non-smooth for the trained model ${\bf X}^\star$ to be sparse. It is worth mentioning that the $k$-th row of ${\bf X}^\star$ (i.e., $({\bm x}_k^\star)^\top$) corresponds to the centroid of the $k$-th true class among $m$ true classes.

In the testing stage, for the given feature vector, denoted as $\boldsymbol{a}^{\prime}\in {\mathbb R}^n$, of an unknown class (whose true class number is given by $k'=\arg\max_k \{b_k^{\prime},k\in[m]\}$), where $b_k^{\prime}$ is the $k$-th element of the true label vector $\bb^{\prime}\in{\mathbb R}^m$.
We compute the following softmax function~\cite{li2022federated}
\begin{align}
z_k = \frac{\exp(({{\bm x}_k^{\star})^{\top} \boldsymbol{a}^{\prime}})}{\sum_{j=1}^{m}\exp({{\bm x}_j^{\top}\boldsymbol{a}^{\prime}})}, \forall k \in [m].
\end{align}
Then the correct decision is made if $\arg\max_k \{z_k\}  = k^{\prime}$. Finally, the overall testing accuracy is obtained as the correct classification rate over the testing dataset.

\subsection{Datasets and Benchmark Algorithms}
\subsubsection{Datasets}
Two benchmark datasets, Adult~\cite{blake1998uci} and MNIST~\cite{website_MNIST}, are considered for performance evaluation. The MNIST dataset consists of 60,000 training samples and 10,000 testing samples, while the Adult dataset consists of 32,561 training samples and 16,281 testing samples. 
To simulate the FL system,  we distribute all training samples across $N = 100$ clients in a non-identical and independently distributed  ({\bf non-i.i.d.})  manner, ensuring that each client possessed partial labels for the data.
\begin{itemize}
  \item  For MNIST dataset for which $(m,n)=(10,785)$, by following the heterogeneous data partition method in~\cite{li2019convergence},  each client is allocated data samples of only four different labels with $|{\cal D}_i|=600$. This leads to a high degree of non-i.i.d. datasets among clients.
  \item For Adult dataset for which $(m,n)=(2,81)$, following \cite{li2020secure}, all the training samples are uniformly distributed among total $N = 100$  clients such that each client only contains data from one class with $|{\cal D}_i|=325$.
\end{itemize}
\subsubsection{Benchmark Algorithms}
 Some state-of-the-art algorithms,
including  FedAvg~\cite{li2020secure}, SCAFFOLD~\cite{karimireddy2020scaffold},  FedProx~\cite{li2020federated} and FedDyn~\cite{durmus2021federated},   are  tested for performance comparison with the proposed  algorithms.

\begin{figure}[t!]
\begin{center}
\resizebox{0.8\linewidth}{!}{\hspace{-0cm}\includegraphics{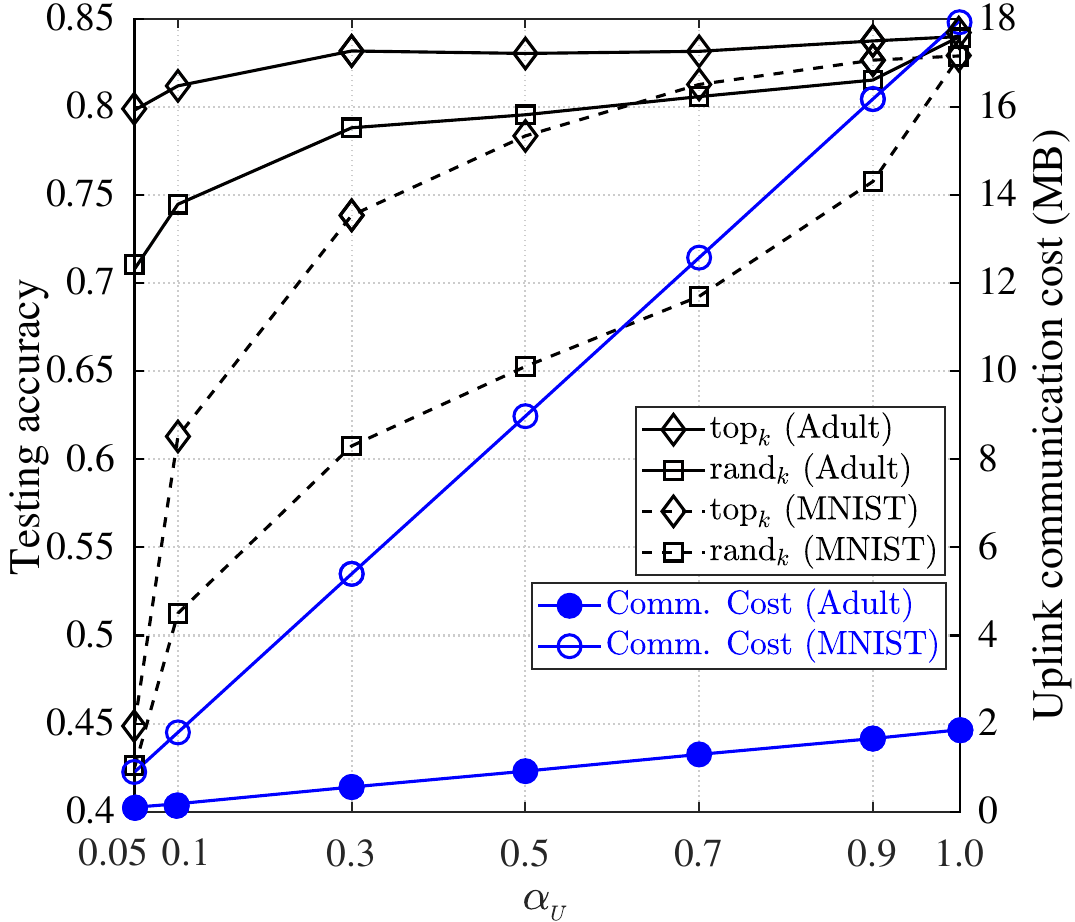}}
\end{center}
\vspace{-0.20cm}
\caption{Testing accuracy and uplink communication costs of the proposed BSDP-FedPDM for  $\alpha_{_D} = 1$ and $\alpha_{_U} \in \{0.05, 0.1, 0.3, 0.5, 0.7, 0.9, 1 \}$.}
\label{MNIST_Adult_top_rand_acc_cost}
\end{figure}

\begin{figure}[t!]
\begin{center}
\resizebox{0.85\linewidth}{!}
{\hspace{-0cm}\includegraphics{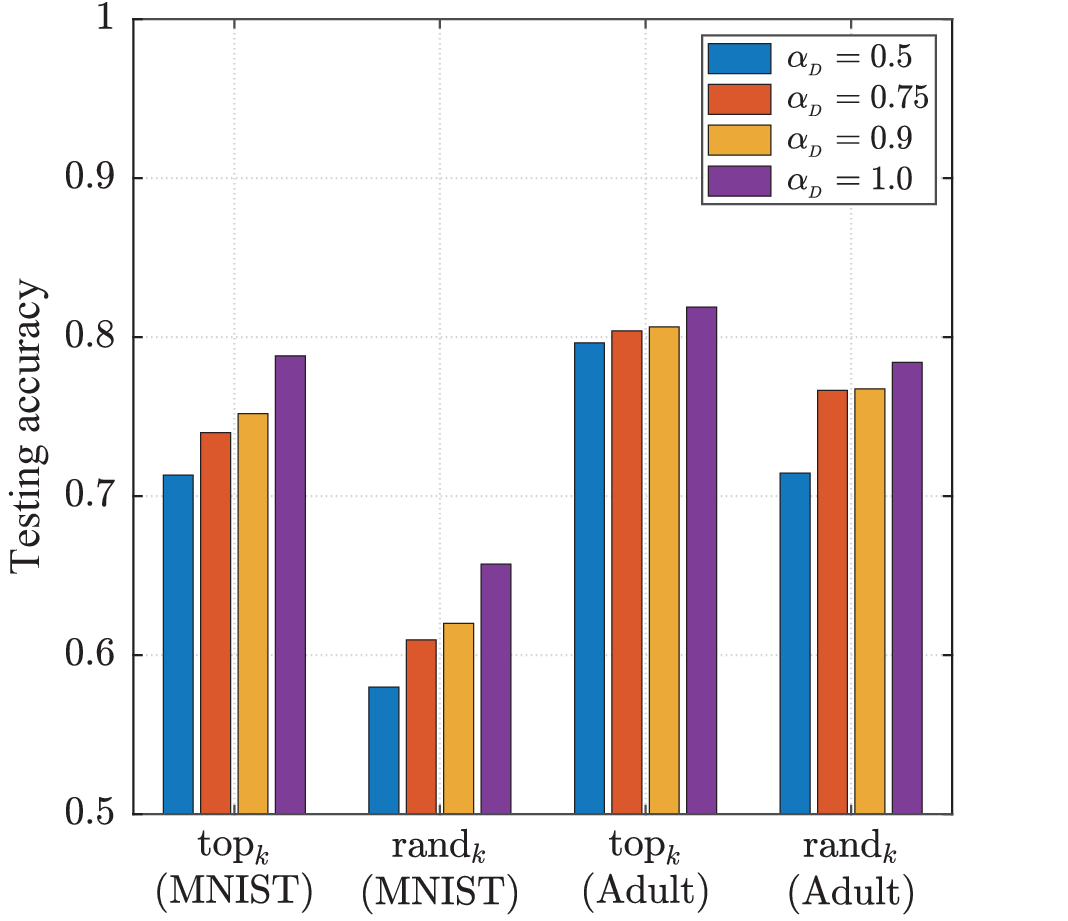}}
\end{center}
\vspace{-0.2cm}
\caption{Testing accuracy of the proposed BSDP-FedPDM for $\alpha_{_D} \in \{0.5, 0.75, 0.9, 1 \}$ and $\alpha_{_U} = 0.5$.}
\label{top_rand_downlink}
\end{figure}

\subsection{Parameter Setting}
For all our experiments, we maintain the following parameter settings: $K=30$, $b=10$, $\rho=10$, $\delta=10^{-4}$, $\nu=10^{-2}$, $\beta = 0.5$, and $\gamma = 0.5$. The choice of the learning rate $\eta^t$  depends on the dataset under consideration.
For MNIST dataset,  the local learning rate is set to $\eta^t=0.04/\sqrt{1+t}$, {while for  Adult dataset,  $\eta^t=0.01/\sqrt{1+t}$.}
For client $i$,  the value of $\epsilon$ used
at each each communication round is set according to \eqref{eqn:total_privacyloss} (cf. Theorem~\ref{Thm:total_privacyloss}), provided that the total privacy loss $\bar\epsilon_i^T, \forall i$ is preassigned. The above system parameters are also the default values used for the experiment, otherwise, they will be clearly specified instead.
\subsection{Experiment Results}
Each experiment is conducted using five different randomly generated initial points. The \emph{uplink communication cost}  refers to the size of the data transmitted from all participated clients to the PS over the entire training process, which  is calculated by $32 k \times T \times K$ bits~\cite{hu2022federated} assuming 32 bits for every real number. Next, let us present the performance of the proposed algorithms under various experimental settings.

\begin{figure}[t!]
\begin{minipage}[b]{0.48\linewidth}
\includegraphics[scale=0.50]{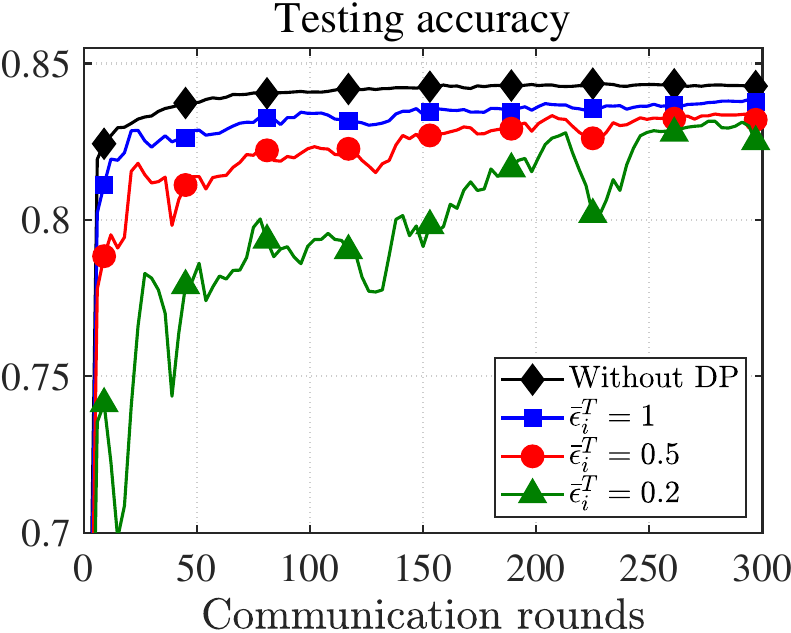}
\centerline{\scriptsize{(a)}}\medskip
\vspace{-0.3cm}
\end{minipage}
\hfill
\begin{minipage}[b]{0.48\linewidth}
\includegraphics[scale=0.50]{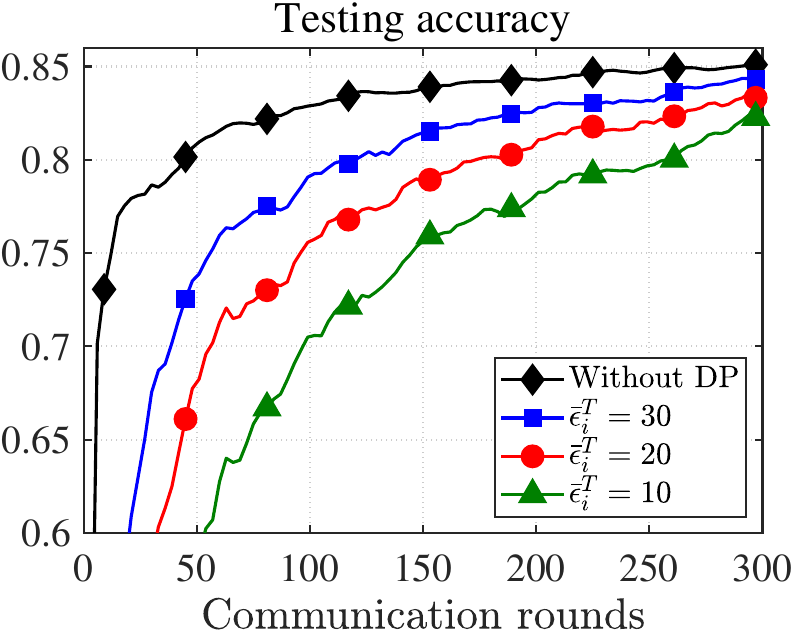}
\centerline{\scriptsize{(b)}}\medskip
\vspace{-0.3cm}
\end{minipage}
\caption{Performance of the proposed DP-FedPDM on (a) Adult dataset for the cases of ``without DP'',  $\bar\epsilon_i^T \in \{0.2,0.5,1\}$ and (b) MNIST dataset for the cases of ``without DP'',   $\bar\epsilon_i^T \in \{10,20,30\}$.}
\label{fig:epsilon_accuracy}
\end{figure}
\begin{figure}[t!]
\begin{minipage}[b]{0.48\linewidth}
\includegraphics[scale=0.50]{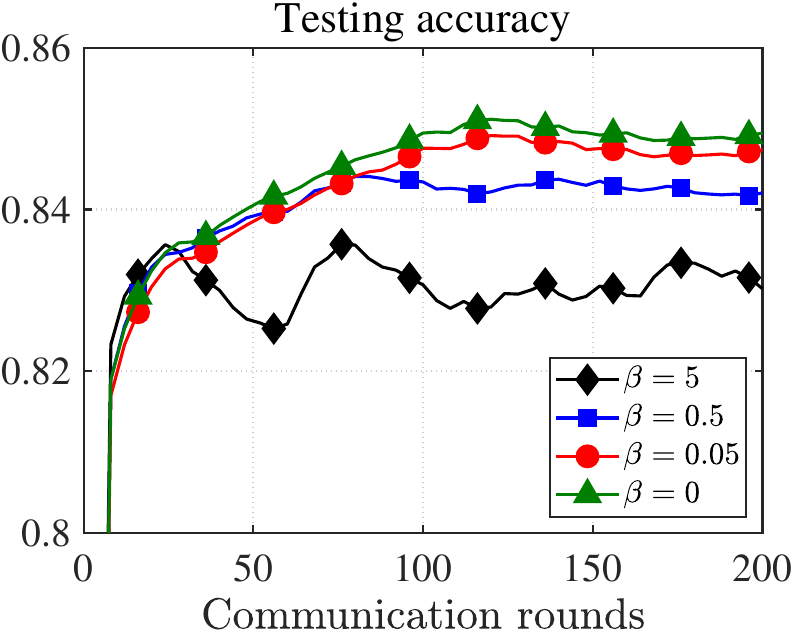}
\centerline{\scriptsize{(a)}}\medskip
\vspace{-0.35cm}
\end{minipage}
\hfill
\begin{minipage}[b]{0.48\linewidth}
\includegraphics[scale=0.50]{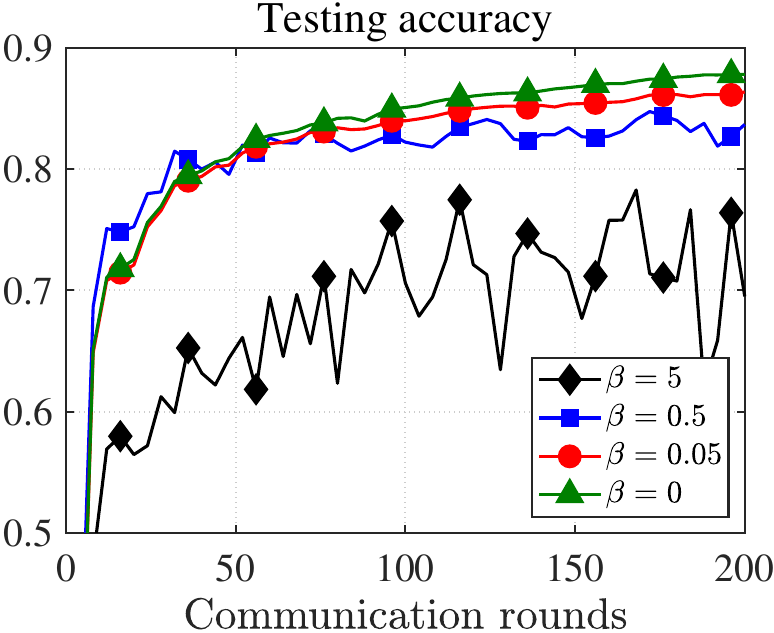}
\centerline{\scriptsize{(b)}}\medskip
\vspace{-0.35cm}
\end{minipage}
\caption{Performance of  DP-FedPDM for $\beta \in \{0, 0.05, 0.5, 5\}$  for (a) Adult dataset, where  $\bar\epsilon_i^T = 0.5$, and (b) MNIST dataset, where $\bar\epsilon_i^T = 20$.}
\vspace{-0.05cm}
\label{fig:convex_nonconvex}
\end{figure}

\begin{figure*}[t!]
\begin{minipage}[b]{0.48\linewidth}
\centering
\includegraphics[scale=0.50]{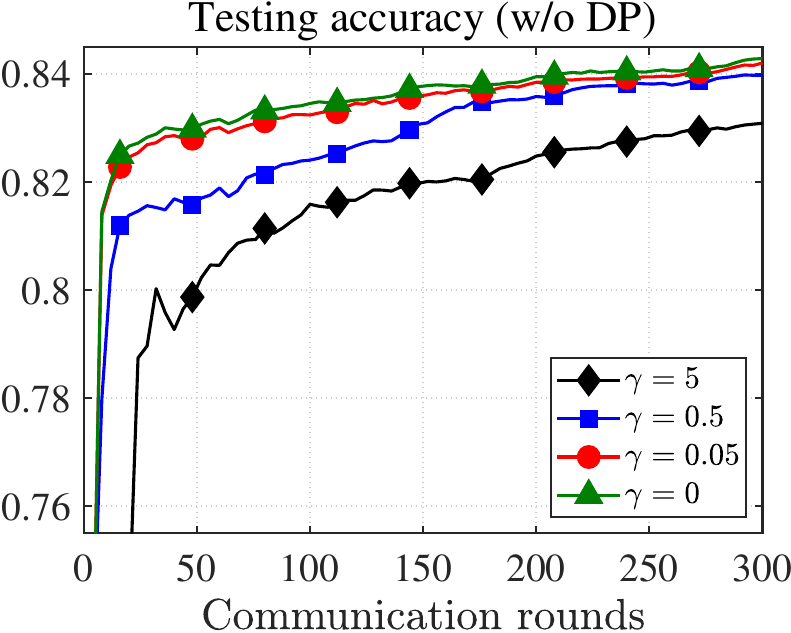}
\centerline{\scriptsize{(a)   }}\medskip
\end{minipage}
\hfill
\begin{minipage}[b]{0.48\linewidth}
\centering
\includegraphics[scale=0.50]{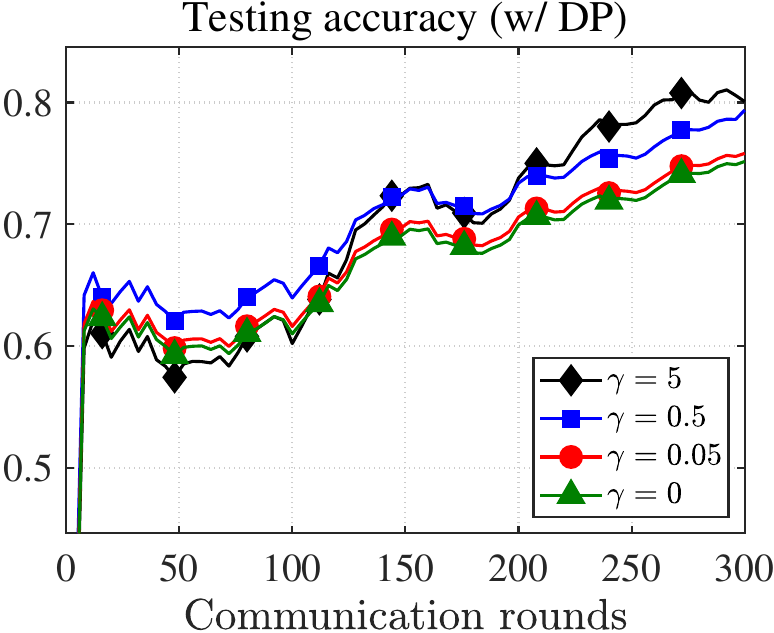}
\centerline{\scriptsize{(b)  }}\medskip
\end{minipage}
\begin{minipage}[b]{0.48\linewidth}
\centering
\includegraphics[scale=0.50]{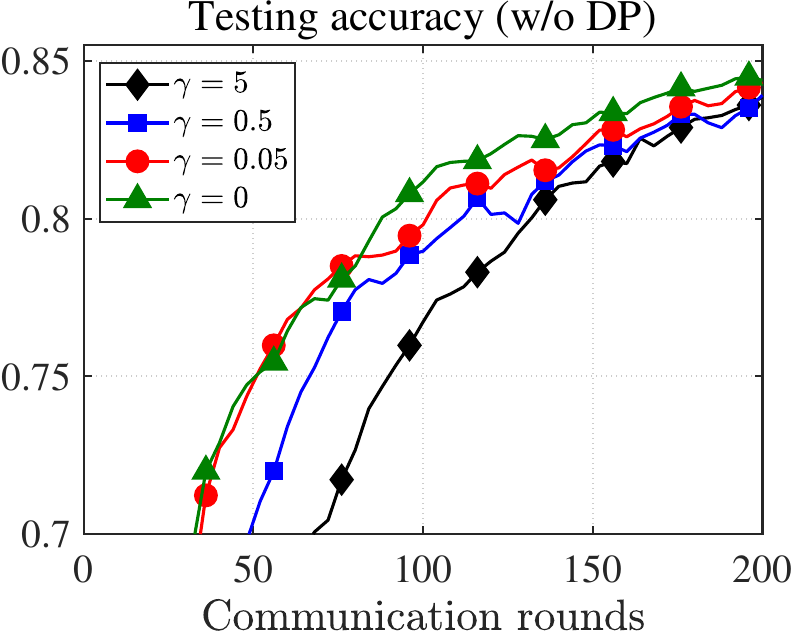}
\centerline{\scriptsize{(c)    }}\medskip
\end{minipage}
\hfill
\begin{minipage}[b]{0.48\linewidth}
\centering
\includegraphics[scale=0.50]{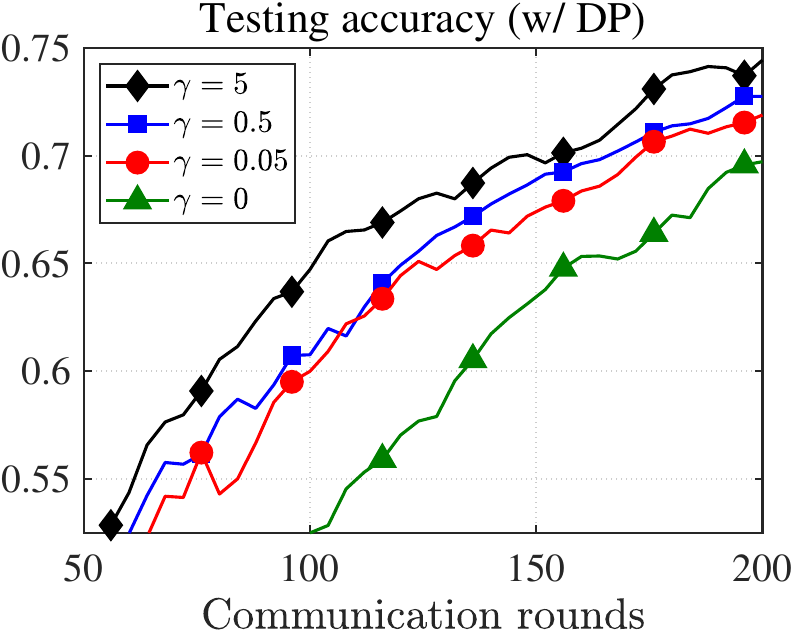}
\centerline{\scriptsize{(d)     }}\medskip
\end{minipage}

\caption{Performance of the proposed DP-FedPDM for the cases of $\gamma \in \{0, 0.05, 0.5, 5\}$ for (a) without DP and (b) $\bar\epsilon_i^T = 0.1$ for Adult dataset, and (c) without DP and (d) $\bar\epsilon_i^T = 5$ for  MNIST dataset.}
\vspace{-0.15cm}
\label{fig:smooth_nonsmooth}
\end{figure*}

\vspace{0.1cm}
\subsubsection{Impact of model sparsification}\label{subsec:impact_of_spar}
We first evaluate the performance (testing accuracy) of the BSDP-FedPDM algorithm for different uplink and downlink compression ratios. In this experiment, we set $T=200$,  $\bar{\epsilon}_i^T = 0.5$ for the Adult dataset and $\bar{\epsilon}_i^T = 20$ for  MNIST dataset.
As shown in Fig. \ref{MNIST_Adult_top_rand_acc_cost},  reducing $\alpha_{_U}$ brings savings in communication cost (i.e., higher communication efficiency) along with  some performance loss, which is consistent with the property (P2)
 of the proposed algorithm. The performance loss of the proposed BSDP-FedPDM is less sensitive to $\alpha_{_U}$ for Adult dataset with $m=2$ than for MNIST dataset with $m=10$. Moreover,
 the $\topk_k$  sparsifier exhibits superior performance over $\randk_k$ sparsifier, especially with significant performance gap for MNIST dataset.


Figure~\ref{top_rand_downlink} illustrates the impact of downlink compression ratio on the performance of BSDP-FedPDM. Notably, MNIST dataset is more sensitive to downlink model compression than Adult dataset.  However, we would like to emphasize if the tradeoff between the communication cost (larger for larger compression ratio)  and the performance (better for larger compression ratio) is application-dependent and/or data-dependent, and controlled by the system operator through the choice of $(\alpha_{_U},\alpha_{_D})$.

\vspace{0.1cm}
\subsubsection{Impact of DP}
Figure~\ref{fig:epsilon_accuracy} shows the experimental results (testing accuracy versus communication round) for the proposed DP-FedPDM for different total privacy loss $\bar{\epsilon}_i^T$. One can observe that DP-FedPDM performs better for larger $\bar{\epsilon}_i^T$ (or weaker privacy protection) with the best performance for the case ``without DP" (no privacy protection), and that its testing accuracy increases with communication round, thus exhibiting the convergence behavior w.r.t. communication round. These results are  consistent with our analyses (cf. Theorems~\ref{Thm:total_privacyloss} and \ref{Thm:Theorem_convergence} and Remark~\ref{remark:remark1}).


\begin{figure*}[t!]
\begin{minipage}[b]{0.48\linewidth}
\centering
\includegraphics[scale=0.50]{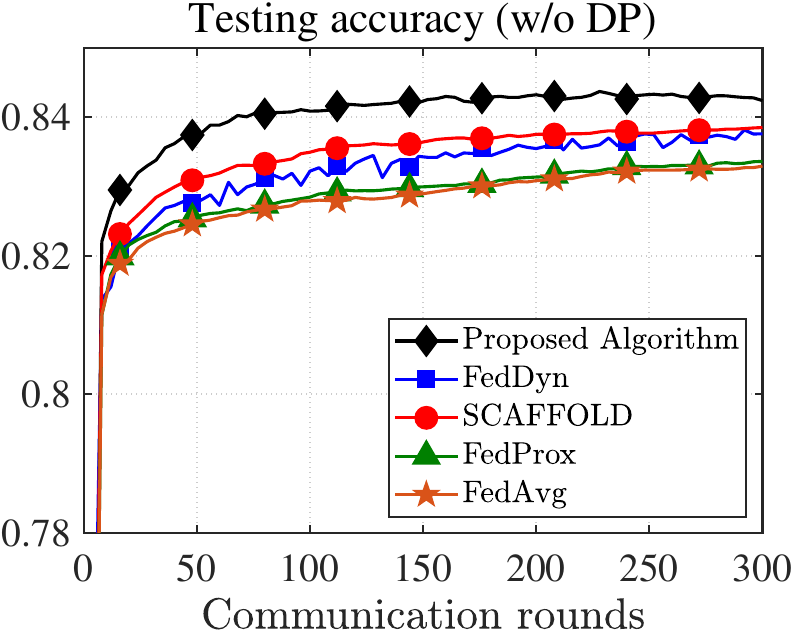}
\centerline{\scriptsize{(a) }}\medskip
\end{minipage}
\hfill
\begin{minipage}[b]{0.48\linewidth}
\centering
\includegraphics[scale=0.50]{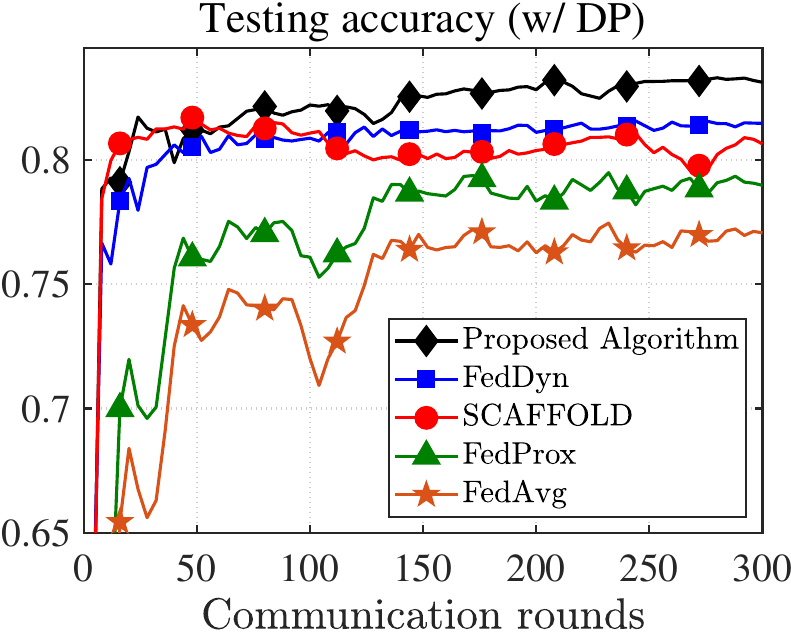}
\centerline{\scriptsize{(b)  }}\medskip
\end{minipage}
\begin{minipage}[b]{0.48\linewidth}
\centering
\includegraphics[scale=0.50]{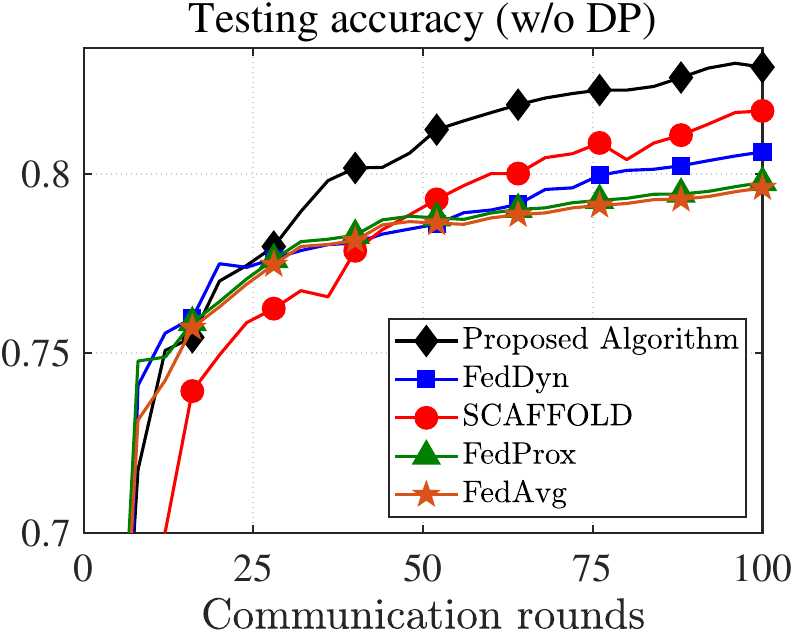}
\centerline{\scriptsize{(c)      }}\medskip
\end{minipage}
\hfill
\begin{minipage}[b]{0.48\linewidth}
\centering
\includegraphics[scale=0.50]{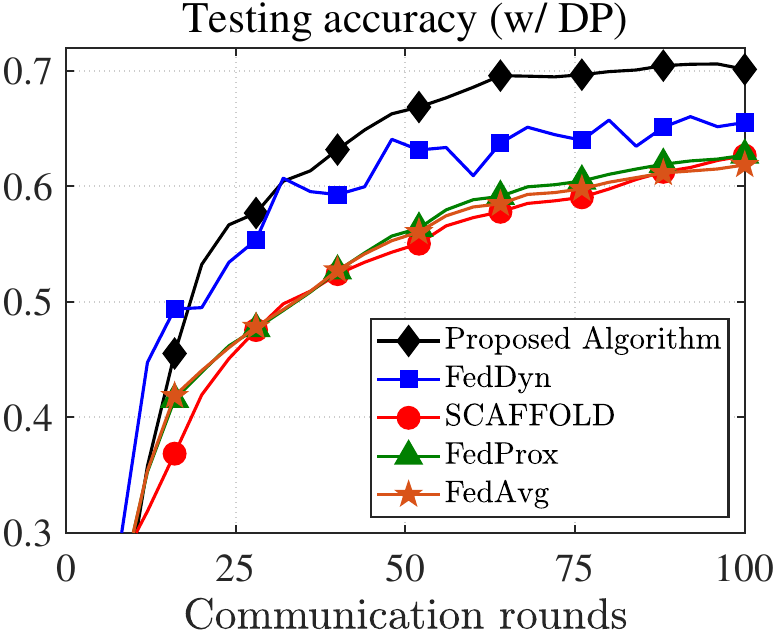}
\centerline{\scriptsize{(d) }}\medskip
\end{minipage}
\caption{Performance comparison between the proposed DP-FedPDM without model sparsification and benchmark algorithms   for the cases of (a) ``without DP" and (b) $\bar\epsilon_i^T = 0.5$ for Adult dataset; and the cases of (c) ``without DP" and (d) $\bar\epsilon_i^T = 10$ for the MNIST dataset.}
\label{model_comparison_epsilon}
\end{figure*}

\begin{figure*}[t!]
\begin{minipage}[b]{0.48\linewidth}
\centering
\includegraphics[scale=0.50]{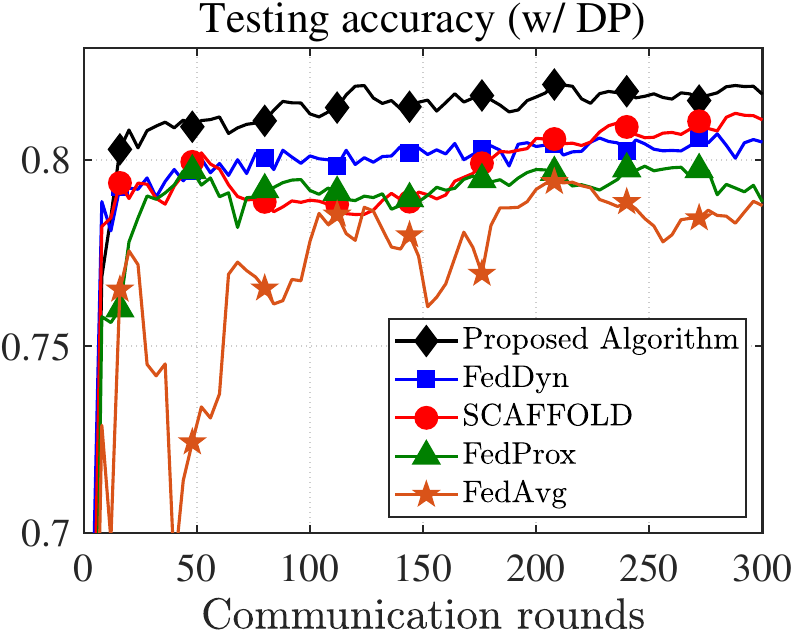}
\centerline{\scriptsize{(a)    }}\medskip
\end{minipage}
\hfill
\begin{minipage}[b]{0.48\linewidth}
\centering
\includegraphics[scale=0.50]{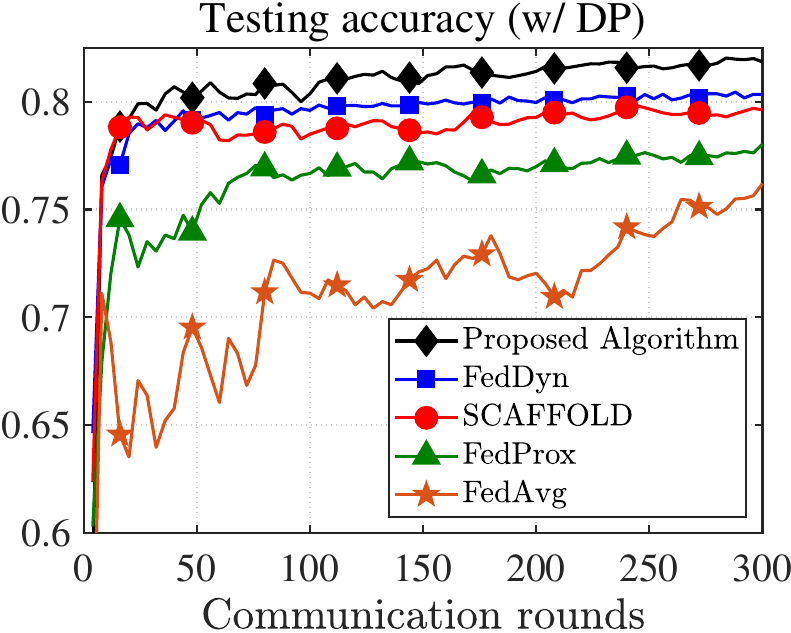}
\centerline{\scriptsize{(b)   }}\medskip
\end{minipage}
\begin{minipage}[b]{0.48\linewidth}
\centering
\includegraphics[scale=0.50]{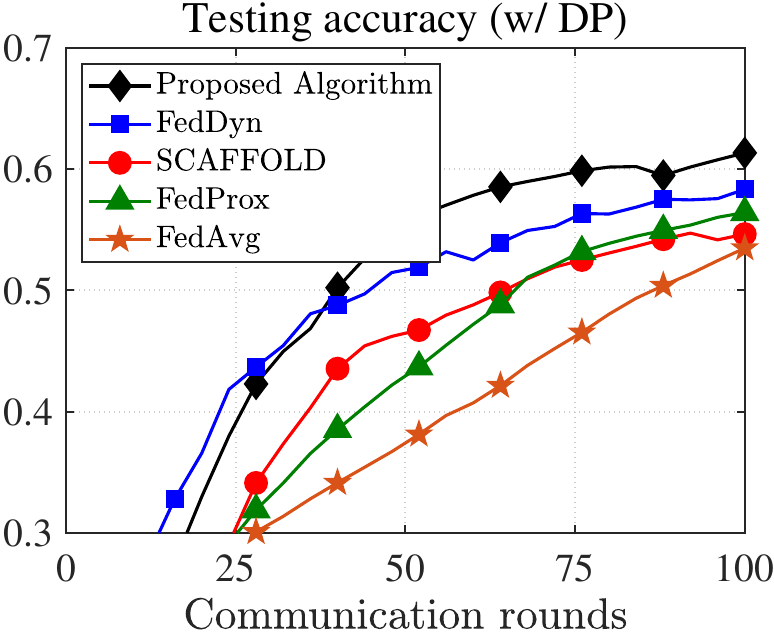}
\centerline{\scriptsize{(c)}}\medskip
\end{minipage}
\hfill
\begin{minipage}[b]{0.48\linewidth}
\centering
\includegraphics[scale=0.50]{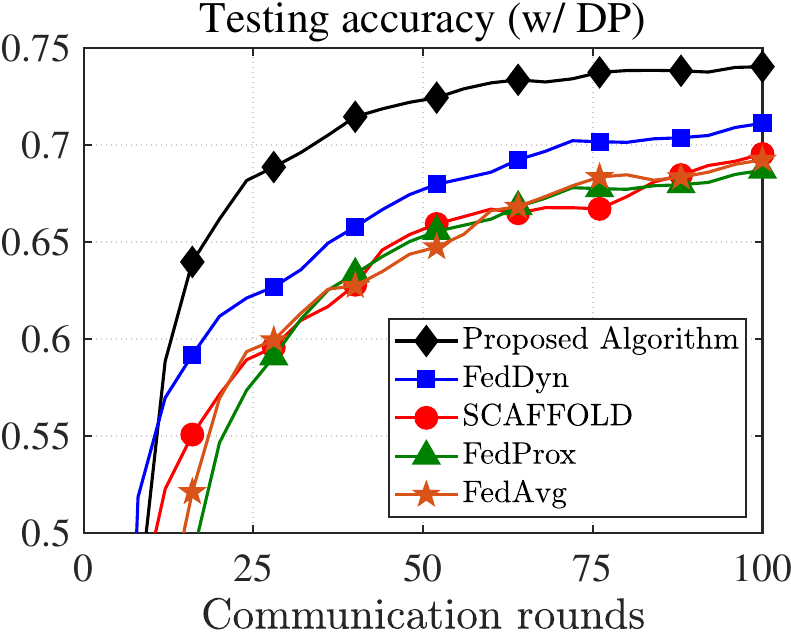}
\centerline{\scriptsize{(d)}}\medskip
\end{minipage}
\caption{Performance comparison between the proposed BSDP-FedPDM and benchmark algorithms for the cases of (a) $(\alpha_{_U},\alpha_{_D}) = (0.1,1)$, and (b) $(\alpha_{_U},\alpha_{_D}) = (0.5,0.75)$ with $\bar\epsilon_i^T =0.5$ for   Adult dataset; and the cases of (c) $(\alpha_{_U},\alpha_{_D}) = (0.1,1)$, and (d) $(\alpha_{_U},\alpha_{_D}) = (0.5,0.75)$ with $\bar\epsilon_i^T =20$ for   MNIST dataset.}
\label{model_comparison_epsilon_spar}
\end{figure*}

\vspace{0.1cm}
\subsubsection{Impact of non-convex loss function}
Figure~\ref{fig:convex_nonconvex} presents the performance of DP-FedPDM for $\beta \in \{0, 0.05, 0.5, 5\}$  while
maintaining $\bar\epsilon_i^T = 0.5$ for Adult dataset and $\bar\epsilon_i^T = 20$ for
MNIST dataset. It can be observed that the smaller the $\beta$, the better its performance simply because the {non-convexity} of the loss function $f_i$ used by client $i$ increases with $\beta$ (cf. \eqref{eq: crossentropy}), implying that the non-convexity of $f_i$ is always a concern in FL system.

\vspace{0.1cm}
\subsubsection{Impact of non-smooth regularizer}\label{subsec:impact_of_nonsmooth}
Figure~\ref{fig:smooth_nonsmooth} illustrates the testing accuracy versus communication round achieved by DP-FedPDM for $\gamma = \{0,0.05,0.5,5\}$. It can be observed from this figure, that the testing accuracy performance is better for the withtout DP (w/o DP) case than for the case with DP (w/ DP), thus consistent with the widely known fact
about the tradeoff between the performance and the privacy protection. Another observation is that the testing accuracy performance for the w/o DP case as shown Figs. \ref{fig:smooth_nonsmooth}(a) and \ref{fig:smooth_nonsmooth}(c) (the case with DP as shown in Figs. \ref{fig:smooth_nonsmooth}(b) and \ref{fig:smooth_nonsmooth}(d)) is better for smaller (larger) $\gamma$, meanwhile  demonstrating that the performance with a suitable $\gamma$ value is robust against DP noise, which is consistent with the properties (P3) and (P4).
\vspace{0.1cm}
\subsubsection{Performance comparison with benchmark algorithms}
Figure~\ref{model_comparison_epsilon} (Figure~\ref{model_comparison_epsilon_spar}) show some comparison
results versus without (with) model sparsification
for the proposed DP-FedPDM (BSDP-FedPDM)
and the above-mentioned four benchmark algorithms. One can see from Fig. \ref{model_comparison_epsilon}, all the algorithms under test incur some performance loss when DP is applied. Nevertheless, we can conclude that
DP-FedPDM outperforms all the other algorithms under test for both with DP and w/o DP cases, in addition to its stronger robustness against the DP noise, as stated in the property (P3).  Moreover, the same observations from Fig. \ref{model_comparison_epsilon} also apply to Fig. \ref{model_comparison_epsilon_spar}. Therefore, the above  conclusions for DP-FedPDM are also true for BSDP-FedPDM.
\section{Conclusions}\label{sec:Conclusions}
We have presented two DP based primal-dual FL algorithms for solving a problem with a non-convex loss function plus a non-smooth regularizer, including DP-FedPDM (without model sparsification) and BSDP-FedPDM (with model sparsification considered) together with their intriguing insightful properties (cf. (P1) through (P4) at the end of Section \ref{subsec:DP-FedPDM}), also suggesting that   the $\topk_k$ sparsifier and   $\ell_1$-norm regularizer is a great match for better learning performance and lower communication cost.  Note that DP-FedPDM is actually a special case of BSDP-FedPDM when  $\alpha_{_U}=\alpha_{_D}=1$. Moreover, we also presented the privacy and convergence analyses for the proposed DP-FedPDM (cf. Theorems~\ref{Thm:total_privacyloss} and \ref{Thm:Theorem_convergence}, Remarks \ref{remark:remark1} and \ref{remark:rmk2}), that can be used as guidelines for the FL algorithm design, especially, an expectant tradeoff between the testing performance and privacy protection level.  Extensive experimental results  on non-i.i.d. real-world data over all the clients  under the practical classification scenario have been provided to demonstrate their efficacy, properties, and analytical results, and much superior performance over some state-of-the-art algorithms. {As a final remark, the proposed algorithms are applicable only when the NCNS loss function can be  mathematically well defined by the training data and the model to be trained, namely, not involving AI-based network model (e.g., Convolutional Neural Networks) that cannot be defined mathematically.
We leave it as a future research for the AI-based model.
}


\numberwithin{equation}{section}
\appendices
\section{Proof of \eqref{eqn:x_0_update_b}}\label{appdixA}
According to \eqref{eq: sadmm-update_5}, we have
\begin{align}
\xb_{0}^{t+1} =& \arg \min\limits_{ \xb_{0}} \mathcal{L} \big( \xb^{t+1}, \xb_{0}, \lambdab^{t+1}\big) \notag \\
\overset{(a)}{=}&\arg \min\limits_{ \xb_{0}}  h(\xb_{0})+\sum_{i=1}^{N} \left( {\lambdab_{i}^{t+1}}^{\top} (\xb_{0} -  \xb_{i}^{t+1})   +  \frac{\rho}{2}  \left\| \xb_{0} -\xb_{i}^{t+1}\right\|^{2} \right)  \notag \\
\overset{(b)}{=}&\arg \min\limits_{ \xb_{0}}  h(\xb_{0})+\frac{\rho}{2} \sum_{i=1}^{N} \Big(\big\|\frac{1}{\rho}\lambdab_{i}^{t+1} \big\|^2 +  \frac{2}{\rho}{\lambdab_{i}^{t+1}}^{\top} (\xb_{0} -  \xb_{i}^{t+1})  +   \left\| \xb_{0} -\xb_{i}^{t+1}\right\|^{2}    \Big)\notag \\
=&\arg \min\limits_{ \xb_{0}}  h(\xb_{0})+\frac{\rho}{2}  \sum_{i=1}^{N} \left(\big\| \xb_{0}-(\xb_{i}^{t+1}-\frac{1}{\rho}\lambdab_i^{t+1})\big\|^2\right) \notag \\
\overset{(c)}{=}& \arg \min\limits_{ \xb_{0}} h(\xb_{0})+  \frac{\rho}{2} \big\| \xb_{0}- \frac{1}{N} \sum_{i=1}^{N}  (\xb_{i}^{t+1}-\frac{1}{\rho}\lambdab_i^{t+1})\big\|^2 \notag \\
\overset{}{=}&{\rm prox}_{\rho^{-1} h}\Big( \frac{1}{N} \sum_{i=1}^{N}  (\xb_{i}^{t+1}-\frac{1}{\rho}\lambdab_i^{t+1}) \Big), \label{eqn:A1}
\end{align}where $(a)$, $(b)$ and $(c)$ hold because   $f_i(\xb_i)$, $\xb_i$, $\lambdab_i, \, \forall i$ are constants w.r.t. $\xb_0$.   $\hfill \blacksquare$

\vspace{0.2cm}
\section{Proof of Theorem \ref{Thm:total_budget}}\label{appdix: proof of Lemma}
Suppose that  $\mathcal{D}_{i}$ and $\mathcal{D}_{i}^{\prime}$ are the neighboring datasets that differ in only one data sample.
For clarity of the following proof,  we introduce the notations $\xb_{i, \mathcal{D}_{i}}$ and $\xb_{i, \mathcal{D}_{i}^{\prime}}$ to represent the updated values of $\xb_{i}$, derived from the datasets $\mathcal{D}_{i}$ and $\mathcal{D}_{i}^{\prime}$, respectively. Similarly, we define $\lambdab_{i, \mathcal{D}_{i}}$ and $\lambdab_{i, \mathcal{D}_{i}^{\prime}}$ as the updated values of $\lambdab_{i}$ from the datasets $\mathcal{D}_{i}$ and $\mathcal{D}_{i}^{\prime}$. Let ${\rm spar}(\cdot)$ denote the sparsifier.
According to \eqref{eqn:global sensitivity_f}, the sensitivity of ${\rm spar}(\yb_{i}^{t+1})$ is given as follows,
\vspace{-0.05cm}
\begin{align} \label{eqn:sensitivity_y}
s_{i,t} =&  \max _{\mathcal{D}_{i}, \mathcal{D}_{i}^{\prime}} \big\| {\rm spar}(\yb_{i, \mathcal{D}_{i}}^{t+1} -  \yb_{i, \mathcal{D}_{i}^{\prime}}^{t+1} ) \big\| \notag \\
\leq& \max _{\mathcal{D}_{i}, \mathcal{D}_{i}^{\prime}} \big\|\yb_{i, \mathcal{D}_{i}}^{t+1} -  \yb_{i, \mathcal{D}_{i}^{\prime}}^{t+1}  \big\| \notag \\
=& \max _{\mathcal{D}_{i}, \mathcal{D}_{i}^{\prime}} \big\|\xb_{i, \mathcal{D}_{i}}^{t+1} -  \xb_{i, \mathcal{D}_{i}^{\prime}}^{t+1} - \frac{1}{\rho} \big(  \lambdab_{i, \mathcal{D}_{i}}^{t+1} -  \lambdab_{i, \mathcal{D}_{i}^{\prime}}^{t+1}\big)  \big\| \notag \\
\overset{(a)}{=}&   \max _{\mathcal{D}_{i}, \mathcal{D}_{i}^{\prime}} \big\|\xb_{i, \mathcal{D}_{i}}^{t+1} -  \xb_{i, \mathcal{D}_{i}^{\prime}}^{t+1} - \frac{1}{\rho} \big(  \lambdab_{i}^{t} -  \lambdab_{i}^{t}\big)   +    \xb_{i, \mathcal{D}_{i}}^{t+1} -  \xb_{i, \mathcal{D}_{i}^{\prime}}^{t+1} + \big(   \xb_{0}^{t} - \xb_{0}^{t} \big)  \big\|  \notag\\
=&  2 \max _{\mathcal{D}_{i}, \mathcal{D}_{i}^{\prime}} \big\|\xb_{i, \mathcal{D}_{i}}^{t+1} -  \xb_{i, \mathcal{D}_{i}^{\prime}}^{t+1}   \big\|,
\end{align}where  $(a)$ holds true due to     $\lambdab_{i}^{t} = \lambdab_{i}^{t-1}- \rho\big( \xb_{i}^{t}-\xb_{0}^{t}\big)$ in \eqref{eq: sadmm-update_3}.  By combining \eqref{eq: sadmm-update}  and \eqref{eqn:sensitivity_y},  we have
\begin{align}
\big\| \xb_{i, \mathcal{D}_{i}}^{t+1} - \xb_{i, \mathcal{D}_{i}^{\prime}}^{t+1} \big\|  = &   \big\|  \xb_{i, \mathcal{D}_{i}}^{t,Q_i^t} - \xb_{i, \mathcal{D}_{i}^{\prime}}^{t,Q_i^t} \big\| \notag \\
=& \Big\| \big( \xb_{i, \mathcal{D}_{i}}^{t,Q_i^t-1} - \eta^t (\nabla f_i (\xb_{i, \mathcal{D}_{i}}^{t,Q_i^t};\Bc_i^t)-\lambdab_{i}^t +\rho (\xb_{i, \mathcal{D}_{i}}^{t,Q_i^t-1}-\xb_{0}^t)) \big) \notag \\
&-\big( \xb_{i, \mathcal{D}_{i}^{\prime}}^{t,Q_i^t-1} - \eta^t (\nabla f_i (\xb_{i, \mathcal{D}_{i}^{\prime}}^{t,Q_i^t};({\Bc_i^t})^{\prime})-\lambdab_{i}^t +\rho (\xb_{i, \mathcal{D}_{i}^{\prime}}^{t,Q_i^t-1}-\xb_{0}^t)) \big) \Big\| \notag \\
\overset{(a)}{\leq}&  \big| 1-\rho \eta^t \big| \big\|  \xb_{i, \mathcal{D}_{i}}^{t,Q_i^t-1} - \xb_{i, \mathcal{D}_{i}^{\prime}}^{t,Q_i^t-1}  \big\| +\eta^t  \big\| \nabla f_i (\xb_{i, \mathcal{D}_{i}}^{t,Q_i^t};\Bc_i^t) - \nabla f_i (\xb_{i, \mathcal{D}_{i}^{\prime}}^{t,Q_i^t};({\Bc_i^t})^{\prime})  \big\|  \notag\\
\overset{(b)}{\leq}&  \big| 1-\rho \eta^t \big|^2  \big\| \xb_{i, \mathcal{D}_{i}}^{t,Q_i^t-2} - \xb_{i, \mathcal{D}_{i}^{\prime}}^{t,Q_i^t-2} \big\| + 2\eta^t G(1+\big|1-\rho \eta^t\big|) \notag \\
\vdots& \notag \\
\leq&  \big| 1-\rho \eta^t \big|^{Q_i^t}  \big\| \xb_{i, \mathcal{D}_{i}}^{t,0} - \xb_{i, \mathcal{D}_{i}^{\prime}}^{t,0} \big\| + 2\eta G \sum_{j=0}^{Q_i^t -1}\big|1-\rho \eta^t\big|^j \notag \\
\overset{\text{$(a)$}}{=}& 2\eta^t G \sum_{j=0}^{Q_i^t -1}\big|1-\rho \eta^t\big|^j, \label{eqn:sensitivity of x}
\end{align}where in $(a)$ holds because  of triangle inequality; $(b)$ follows due to Assumption~\ref{Ass: Assumption3} and $(c)$ holds true since $\xb_{i, \mathcal{D}_{i}}^{t,0} = \xb_{i, \mathcal{D}_{i}}^{t-1,Q_i^t}$ and $\xb_{i, \mathcal{D}_{i}}^{t-1,Q_i^t}=\xb_{i, \mathcal{D}_{i}^\prime}^{t-1,Q_i^t}$;  By combining \eqref{eqn:sensitivity_y} and \eqref{eqn:sensitivity of x}, {we end up with $s_{i,t}$ given by \eqref{eqn:sensitivity}, thus   completing the proof of  Theorem \ref{Thm:total_budget}. $\hfill\blacksquare$}			

\section{Proof of Theorem \ref{Thm:Theorem_convergence}}\label{appenx:C}
{Prior to the proof of Theorem 3, we need the following two lemmas, and their proofs are given in Appendix~\ref{sec:appendix_D}. }
\vspace{0.3cm}
\begin{Lemma}\label{Lemma:decreasing_AL_}
Suppose that Assumptions \ref{Ass: Assumption1}-\ref{Ass: Assumption2} hold and $2\sqrt{5}-4\leq L\leq \rho/4$. Then,
\begin{align}   \label{Lemma:decreasing_AL}
&\mathbb{E}\Big[\mathcal{L}( \{\xb_{i}^{t+1}\}, \xb_{0}^{t+1}, \{\lambdab_{i}^{t+1}\} ) - \mathcal{L}( \{\xb_{i}^{t}\}, \xb_{0}^{t}, \{\lambdab_{i}^{t}\} )  \Big] -2\left[ \big( \frac{8}{\rho}+\frac{L}{4L-2}\big)\nu + \frac{16}{\rho } \phi^2\right]\notag\\
\leq&  - \underbrace{\frac{\rho N}{2}}_{B_1} \| \xb_{0}^{t} - \xb_{0}^{t+1}\|^2  +   \sum_{i \in \mathcal{S}_{t}} \Big[ -\big( \underbrace{\frac{\rho}{2}-\frac{4L^2}{\rho} -L \big)}_{B_2}\|\xb_{i}^t-\xb_{i}^{t+1} \|^2   - \underbrace{\big( \frac{8}{\rho}+\frac{L}{4L-2}\big)}_{B_3}\nu -\underbrace{\frac{16}{\rho }}_{B_4} \phi^2\Big].
\end{align}
\end{Lemma}
\noindent \textit{Proof:}  See Appendix  \ref{sec: proof_of_Lemma2}.

\begin{Lemma} \label{Lemma:Expected_P_}
Suppose that Assumptions \ref{Ass: Assumption1}-\ref{Ass: Assumption2} hold. Then,
\begin{align} \label{Lemma:Expected_P}
&\,\mathbb{E}\Big[P(\{\xb_{i}^{t}\}, \xb_{0}^{t}, \{\lambdab_{i}^{t}\})  \Big] \notag\\
\leq&~ \underbrace{2 }_{B_5}\| \mathbf{x}_{0}^{t} -\xb_{0}^{t+1}\|^2
+  \sum_{i\in {\cal S}_t} \Big\{ \underbrace{( 1+\frac{4\rho}{N})\frac{16}{\rho^2 }}_{\triangleq B_6} \phi^2 \notag\\
& + \underbrace{\big[(L^2+\rho^2) +\frac{4}{N}(\rho+\frac{4L^2}{\rho})+\frac{4L^2}{\rho^2} \big]}_{\triangleq B_7} \| \xb_{i}^{t+1} -\xb_i^t\|^2	  \notag\\
&+ \underbrace{\big(\frac{L^2}{4L-2}+(1+\frac{4\rho}{N})\frac{8}{\rho^2}\big)}_{\triangleq B_8}  \nu \Big\} + 8\rho \sigma^2.
\end{align}
\end{Lemma}
\noindent \textit{Proof:}  See Appendix  \ref{sec: proof_of_Lemma3}.
\vspace{0.2cm}
{Now let us present the proof.} With $P(\{\xb_{i}^{t}\}, \xb_{0}^{t}, \{\lambdab_{i}^{t}\})$ defined in \eqref{eqn:criterion} and Lemma \ref{Lemma:decreasing_AL_}, Lemma \ref{Lemma:Expected_P_} above, we have the following inferences:
\begin{align}
&  \frac{1}{T} \sum_{t=0}^{T-1} \mathbb{E}\big[P(\{\xb_{i}^{t}\}, \xb_{0}^{t}, \{\lambdab_{i}^{t}\}) \big] \notag\\
\overset{(a)}{\leq}&\,\frac{1}{T} \sum_{t=0}^{T-1}\theta_2\Big\{\| \mathbf{x}_{0}^{t} -\xb_{0}^{t+1}\|^2 + \sum_{i\in {\cal S}_t} \big( \| \xb_{i}^{t+1} -\xb_i^t\|^2   +\nu+\phi^2 \big) \Big\} + 8\rho \sigma^2\notag\\
\overset{}{=}&\, \frac{1}{T} \sum_{t=0}^{T-1}\frac{\theta_1\theta_2}{\theta_1}\Big\{\| \mathbf{x}_{0}^{t} -\xb_{0}^{t+1}\|^2
+  \sum_{i\in {\cal S}_t}\big( \| \xb_{i}^{t+1} -\xb_i^t\|^2   +   \nu+\phi^2 \big) \Big\}
+ 8\rho \sigma^2\notag\\
\overset{(b)}{\leq} &\,\frac{1}{T} \sum_{t=0}^{T-1} \frac{\theta_2}{\theta_1}\Big\{ \mathbb{E}\big[\mathcal{L}( \{\xb_{i}^{t}\}, \xb_{0}^{t}, \{\lambdab_{i}^{t}\} ) -\mathcal{L}( \{\xb_{i}^{t+1}\}, \xb_{0}^{t+1}, \{\lambdab_{i}^{t+1}\} ) \big]\Big\}  +C_1\nu+C_2  \phi^2   + C_3 \sigma^2\notag \\
\overset{(c)}{\leq} &   C_0 \frac{\big( \mathcal{L}( \{\xb_{i}^{0}\}, \xb_{0}^{0}, \{\lambdab_{i}^{0}\} ) - \underline{f} \big)}{T} +C_1\nu+C_2  \phi^2   + C_3 \sigma^2, \label{eqn:proof_thm3}
\end{align}which is exactly \eqref{eqn:Thm1}, where $(a)$, $(b)$, $(c)$ hold due to Lemma~\ref{Lemma:Expected_P_}, Lemma~\ref{Lemma:decreasing_AL_}, and Assumption \ref{Ass: Assumption1}, respectively; and
\begin{align}
&\theta_1=\min\{B_1,B_2,B_3,B_4\},  \theta_2 =\max\{B_5,B_6,B_7,B_8\}  \label{eqn:C4_1},\\
&C_0= \frac{\theta_2}{\theta_1}, C_1=\frac{2\theta_2 B_3}{\theta_1}, C_2=\frac{2\theta_2 B_4}{\theta_1},  C_3 = 8\rho. \label{eqn:C4_2}
\end{align}$\hfill \blacksquare$

\vspace{-0.3cm}
\section{Proof of Key Lemmas}\label{sec:appendix_D}
\subsection{Proof of Lemma \ref{Lemma:decreasing_AL_}}\label{sec: proof_of_Lemma2}
We divide the term $\big[\mathcal{L}( \{\xb_{i}^{t+1}\}, \xb_{0}^{t+1}, \{\lambdab_{i}^{t+1}\} ) - \mathcal{L}( \{\xb_{i}^{t}\}, \xb_{0}^{t}, \{\lambdab_{i}^{t}\} )\big]$ into the sum of three parts:			
\begin{align} \label{Lemma1_1}
&\mathcal{L}( \{\xb_{i}^{t+1}\}, \xb_{0}^{t+1}, \{\lambdab_{i}^{t+1}\} ) - \mathcal{L}( \{\xb_{i}^{t}\}, \xb_{0}^{t}, \{\lambdab_{i}^{t}\} ) \notag \\
=&  \underbrace{\mathcal{L}( \{\xb_{i}^{t+1}\}, \xb_{0}^{t+1}, \{\lambdab_{i}^{t+1}\} ) - \mathcal{L}( \{\xb_{i}^{t+1}\}, \xb_{0}^{t}, \{\lambdab_{i}^{t+1}\} )}_{\triangleq A_4} \notag\\
&+  \underbrace{\mathcal{L}( \{\xb_{i}^{t+1}\}, \xb_{0}^{t}, \{\lambdab_{i}^{t+1}\} ) - \mathcal{L}( \{\xb_{i}^{t+1}\}, \xb_{0}^{t},  \{\lambdab_{i}^{t}\} )}_{\triangleq A_5} \notag \\
&+  \underbrace{\mathcal{L}( \{\xb_{i}^{t+1}\}, \xb_{0}^{t}, \{\lambdab_{i}^{t}\} ) - \mathcal{L}( \{\xb_{i}^{t}\}, \xb_{0}^{t}, \{\lambdab_{i}^{t}\} )}_{\triangleq A_6}.			
\end{align}
The $A_4$ term in \eqref{Lemma1_1} can be bounded by
\begin{align}  \label{Lemma1_2}
&\mathcal{L}( \{\xb_{i}^{t+1}\}, \xb_{0}^{t+1}, \{\lambdab_{i}^{t+1}\} ) - \mathcal{L}( \{\xb_{i}^{t+1}\}, \xb_{0}^{t}, \{\lambdab_{i}^{t+1}\} ) \notag \\
\overset{\text{$(a)$}}{\leq} &  {(-\psi_{\xb_{0}}^{t+1})}^{\top} (\xb_{0}^{t} - \xb_{0}^{t+1})
-\frac{\rho N}{2} \| \xb_{0}^{t} - \xb_{0}^{t+1}\|^2 \notag  \\
\overset{\text{$(b)$}}{\leq} &- \frac{\rho N}{2} \| \xb_{0}^{t} - \xb_{0}^{t+1}\|^2 ,
\end{align}
where  $(a)$ holds due to the fact that $\mathcal{L}( {\xb_{i}^{t+1}}, \xb_{0}, {\lambdab_{i}^{t+1}} )$ is strongly convex with respect to $\xb_{0}$, with modulus $\rho N$, and $\psi_{\xb_{0}}^{t+1}$ is a subgradient of $ \mathcal{L}( {\xb_{i}^{t+1}}, \xb_{0}^{t+1}, {\lambdab_{i}^{t+1}})$; $(b)$ holds because $\xb_{0}^{t+1}$ is the optimal solution of the strongly convex function $\mathcal{L}( {\xb_{i}^{t+1}}, \xb_{0}, {\lambdab_{i}^{t+1}} )$, which implies ${\psi_{\xb_{0}}^{t+1}}^{\top}( \xb_{0}^{t} - \xb_{0}^{t+1} ) \geq 0$.


The $A_5$ term in \eqref{Lemma1_1} can be bounded by
\begin{align}
&\mathcal{L}( \{\xb_{i}^{t+1}\}, \xb_{0}^{t}, \{\lambdab_{i}^{t+1}\} ) - \mathcal{L}( \{\xb_{i}^{t+1}\}, \xb_{0}^{t}, \{\lambdab_{i}^{t}\} )  \notag \\
=& \sum_{i \in \mathcal{S}_{t}}  (\lambdab_i^{t+1} - \lambdab_{i}^t)^{\top}( \xb_{0}^t-\xb_{i}^{t+1})   \notag\\
\overset{\text{$(a)$}}{=}& \sum_{i \in \mathcal{S}_{t}} \frac{1}{\rho} \big\| \lambdab_i^{t+1} - \lambdab_{i}^t \big\|^2 \notag \\
\overset{\text{$(b)$}}{=}& \sum_{i \in \mathcal{S}_{t}} \frac{1}{\rho}\big\| (\nabla f_i(\xb_{i}^{t+1};\Bc_i^{t+1})-e_{i}^{t+1}) - (\nabla f_i(\xb_{i}^{t};\Bc_i^{t})-e_{i}^{t}) \big\|^2  \notag\\
\overset{\text{$(c)$}}{\leq}& \sum_{i \in \mathcal{S}_{t}} \frac{2}{\rho} \big(\big\|( \nabla f_i(\xb_{i}^{t+1})+\varepsilon_{i}^{t+1}) - (\nabla f_i(\xb_{i}^{t})+\varepsilon_i^t) \big\|^2 +   4\|e_i\|^2\big) \notag \\
\overset{\text{$(d)$}}{\leq}& \sum_{i \in \mathcal{S}_{t}} \frac{2}{\rho} \big[ 2\big(\| \nabla f_i(\xb_{i}^{t+1})- \nabla f_i(\xb_{i}^{t})\|^2 + 4\|\varepsilon_{i} \|^2 \big)+  4\|e_i\|^2 \big]  \notag\\
\overset{\text{$(e)$}}{\leq}& \sum_{i \in \mathcal{S}_{t}} \frac{4L^2}{\rho} \|\xb_i^{t+1}-\xb_i^t\|^2 +\frac{16}{\rho}\|\varepsilon_i\|^2 + \frac{8}{\rho}\|e_i\|^2 ,  \label{Lemma1_3}
\end{align}
where $\varepsilon_i^{t+1} \triangleq \nabla f_i(\xb_{i}^{t+1};\Bc_i^{t+1})-\nabla f_i(\xb_{i}^{t+1})$ and $\|e_i\|={\max}_t\{\|e_i^t\|\}$,  $(a)$ follows because of \eqref{eq: sadmm-update_3};   $(b)$ holds since \eqref{eq: sadmm-update_1} and \eqref{eq: sadmm-update_2}, and $\mathbb{E}[\|e_i^{t+1} \|^2] \leq \nu $ with
\begin{equation}
\begin{aligned} \label{eqn:Lemma1_4}
e_i^{t+1} = \nabla f_i (\xb_{i}^{t+1};\Bc_i^{t+1}) - \lambdab_{i}^t + \rho (\xb_{i}^{t+1}-\xb_{0}^t).
\end{aligned}
\end{equation}
By combining \eqref{eqn:Lemma1_4} and \eqref{eq: sadmm-update_3}, we have  $\lambdab_{i}^{t+1}=\nabla f_i (\xb_{i}^{t+1};\Bc_i^{t+1}) - e_i^{t+1}$. In  $(c)$, we utilize the fact that $\|\ab\|^2 \leq 2(\|\bb\|^2+\|\cb\|^2)$. Similarly, in $(d)$, we apply the same reasoning as in $(c)$ and set $\|\varepsilon_i\|=\max_t\{\| \varepsilon_i^t\|\}$. For $(e)$, we use the fact that $f_i(\cdot)$ is $L$-smooth.

The $A_6$ term in \eqref{Lemma1_1} can be bounded by
\begin{align}  \label{Lemma1_5}
&~\mathcal{L}( \{\xb_{i}^{t+1}\}, \xb_{0}^{t}, \{\lambdab_{i}^{t}\} ) - \mathcal{L}( \{\xb_{i}^{t}\}, \xb_{0}^{t}, \{\lambdab_{i}^{t}\} )   \notag\\
=&\sum_{i \in \mathcal{S}_{t}} \big[f_i(\xb_{i}^{t+1}) - f_i(\xb_{i}^t) +  {\lambdab_i^{t}}^{\top}  (\xb_{i}^t-\xb_{i}^{t+1}) +\frac{\rho}{2} (\| \xb_{i}^{t+1} - \xb_{0}^t\|^2 -\|\xb_{i}^t-\xb_{0}^t\|^2 ) \big] \notag\\
\overset{\text{$(a)$}}{\leq}&\sum_{i \in \mathcal{S}_{t}} (-\nabla f_i(\xb_{i}^{t+1}))^{\top}(\xb_{i}^t-\xb_{i}^{t+1} ) + \frac{L}{2}\|\xb_{i}^t - \xb_{i}^{t+1} \|^2 \notag \\
&+ {\lambdab_i^{t}}^{\top}  (\xb_{i}^t-\xb_{i}^{t+1})  +\frac{\rho}{2} (\| \xb_{i}^{t+1} - \xb_{0}^t\|^2 -\|\xb_{i}^t-\xb_{0}^t\|^2 )  \notag\\
\overset{\text{$(b)$}}{=}& 	\sum_{i \in \mathcal{S}_{t}} (\lambdab_{i}^t - \nabla f_i(\xb_{i}^{t+1}))^{\top} (\xb_{i}^t-\xb_{i}^{t+1}  )	+ \frac{L}{2}\|\xb_{i}^t - \xb_{i}^{t+1} \|^2  + \frac{\rho}{2}  (-\xb_{i}^{t+1} -\xb_{i}^t + 2\xb_{0}^t)^{\top} (\xb_{i}^t - \xb_{i}^{t+1})  \notag\\
=&  \sum_{i \in \mathcal{S}_{t}} ( \lambdab_{i}^t - \nabla f_i(\xb_{i}^{t+1}) - \rho(\xb_{i}^{t+1}-\xb_0^t))^{\top} (\xb_{i}^t-\xb_{i}^{t+1})  + \frac{L}{2}\|\xb_{i}^t - \xb_{i}^{t+1} \|^2 - \frac{\rho}{2} \|\xb_{i}^t - \xb_{i}^{t+1}\|^2  \notag\\
\overset{\text{$(c)$}}{\leq}&  \sum_{i \in \mathcal{S}_{t}}  \frac{1}{2L}\| \nabla_{\xb_{i}} \mathcal{L} (\{\xb_{i}^{t+1}\},\xb_{0}^t,\{\lambdab_{i}^t\} \|^2 + \big(L-\frac{\rho}{2}\big)\| \xb_{i}^t - \xb_{i}^{t+1}\|^2 \notag \\
\overset{\text{$(d)$}}{\leq}& \sum_{i \in \mathcal{S}_{t}} \frac{L}{4L-2} \nu +  \big(L-\frac{\rho}{2}\big)\| \xb_{i}^t - \xb_{i}^{t+1}\|^2,
\end{align}where  $(a)$ holds because the fact that $-f_i(\cdot)$ is $L$-smooth; in $(b)$ we apply the fact that $\|\ab\|^2-\|\bb\|^2=( -\ab-\bb)^{\top}(-\ab+\bb )$; in $(c)$ we invoke the fact that $ \xb^{\top}\yb  \leq \frac{1}{2L}\|\xb\|^2 +\frac{L}{2} \|\yb\|^2$; and  $(d)$ holds because of the reasoning as follows:
\begin{align} \label{2022-5-5-2126}
&\, \big\| \nabla_{\xb_{i}} \mathcal{L} (\{\xb_{i}^{t+1}\},\xb_{0}^t,\{\lambdab_{i}^t\} \big\|^2 \notag \\
=&  \mathbb{E} \Big[ \big(\nabla_{\xb_{i}} \mathcal{L} (\{\xb_{i}\},\xb_{0}^t,\{\lambdab_{i}^t\})\big)^{\top} \big(\nabla f_i(\xb_{i}^{t+1};\Bc_i^{t+1}) -\lambdab_{i}^t+\rho(\xb_{i}^{t+1}-\xb_{0}^t)\big) \Big] \notag \\
\overset{\text{$(a)$}}{\leq}& \, \mathbb{E}\big[ \frac{1}{2L} \|\nabla_{\xb_{i}} \mathcal{L} (\{\xb_{i}^{t+1}\},\xb_{0}^t,\{\lambdab_{i}^t\} \|^2 +\frac{L}{2}\|e_i^{t+1}\|^2 \big] \notag \\
\overset{\text{$(b)$}}{\leq}& \,  \frac{1}{2L} \big\|\nabla_{\xb_{i}} \mathcal{L} (\{\xb_{i}^{t+1}\},\xb_{0}^t,\{\lambdab_{i}^t\} \big\|^2 +\frac{L}{2} \nu,
\end{align}
where  $(a)$ holds due to the fact that $\xb^{\top}\yb \leq \frac{1}{2L}\|\xb\|^2 +\frac{L}{2} \|\yb\|^2$ and  \eqref{eqn:Lemma1_4}; in $(b)$ we use the update rule in \eqref{eqn:localacc}. Then, we can rewrite   \eqref{2022-5-5-2126} as follows:
\begin{align} \label{2022-5-5-2200}
\big\| \nabla_{\xb_{i}} \mathcal{L} (\{\xb_{i}^{t+1}\},\xb_{0}^t,\{\lambdab_{i}^t\} \big\|^2
\leq \frac{1}{(1-\frac{1}{2L})}\frac{L}{2} \nu =\frac{L^2}{2L-1} \nu.
\end{align}
By combining the results of \eqref{Lemma1_2}, \eqref{Lemma1_3}  and \eqref{Lemma1_5} and   assuming that $\big( \frac{4L^2}{\rho} +L-\frac{\rho}{2} \big) \leq 0$ and $\frac{L^2}{2L-1} \geq 0$, i.e., $2\sqrt{5}-4\leq L\leq \rho/4$, we have
\begin{align} \notag
&\mathbb{E}\Big[\mathcal{L}( \{\xb_{i}^{t+1}\}, \xb_{0}^{t+1}, \{\lambdab_{i}^{t+1}\} ) - \mathcal{L}( \{\xb_{i}^{t}\}, \xb_{0}^{t}, \{\lambdab_{i}^{t}\} )  \Big]\\
\leq&  - \frac{\rho N}{2} \| \xb_{0}^{t} - \xb_{0}^{t+1}\|^2 +   \sum_{i \in \mathcal{S}_{t+1}} \big[ \big( \frac{4L^2}{\rho} +L-\frac{\rho}{2} \big)\|\xb_{i}^t-\xb_{i}^{t+1} \|^2 +\big( \frac{8}{\rho}+\frac{L}{4L-2}\big)\nu
+\frac{16}{\rho} \phi^2\big].
\end{align}
Thus, we complete the proof.   $\hfill \blacksquare$

\subsection{Proof of Lemma \ref{Lemma:Expected_P_}}\label{sec: proof_of_Lemma3}
We first revisit the definition of $P(\{\xb_{i}^{t}\}, \xb_{0}^{t}, \{\lambdab_{i}^{t}\})$ in \eqref{eqn:criterion} as follows,

\vspace{-0.6cm}
\begin{align}\label{eqn:criterion_appendix}
P(\{\xb_{i}^{t}\}, \xb_{0}^{t}, \{\lambdab_{i}^{t}\})  \triangleq & \sum_{j=1}^{N}\Big[ \underbrace{\big\| \nabla_{\xb_{j}} \mathcal{L}( \{\xb_{i}^{t}\}, \xb_{0}^{t}, \{\lambdab_{i}^{t}\} )\big\|^2}_{\triangleq E_1}  \notag \\
&+ \underbrace{\big\| \nabla_{\lambdab_{j}}\mathcal{L}( \{\xb_{i}^{t}\}, \xb_{0}^{t}, \{\lambdab_{i}^{t}\} )  \big\|^2  \Big]}_{\triangleq E_2} \notag\\
&+ \underbrace{\big\| \nabla_{\xb_{0}}\mathcal{L}( \{\xb_{i}^{t}\}, \xb_{0}^{t}, \{\lambdab_{i}^{t}\} ) \big\|^2}_{\triangleq E_3}.
\end{align}
Next, we aim to bound $E_1$, $E_2$, and $E_3$ respectively. Specifically, according to \eqref{2022-5-5-2200}, $E_1$ can be bounded by
\begin{align} \label{Lemma2_1}
&~\big\| \nabla_{\xb_{j}} \mathcal{L}( \{\xb_{i}^{t}\}, \xb_{0}^{t}, \{\lambdab_{i}^{t}\} )\big\|^2 \notag \\
\overset{\text{$(a)$}}{\leq}& \, \frac{1}{2} \big(\big\| \nabla_{\xb_{j}} \mathcal{L}( \{\xb_{i}^{t}\}, \xb_{0}^{t}, \{\lambdab_{i}^{t}\} ) - \nabla_{\xb_{j}} \mathcal{L}( \{\xb_{i}^{t+1}\}, \xb_{0}^{t}, \{\lambdab_{i}^{t}\} )\big\|^2 + \big\|\nabla_{\xb_{j}} \mathcal{L}( \{\xb_{i}^{t+1}\}, \xb_{0}^{t}, \{\lambdab_{i}^{t}\} )\big\|^2 \big) \notag \\
\overset{\text{$(b)$}}{\leq}& \, \frac{1}{2} \big(\big\| (\nabla f_j(\xb_{j}^t) - \lambdab_{j}^t +\rho (\xb_{j}^t -\xb_{0}^t) ) - (\nabla f_j(\xb_{j}^{t+1}) - \lambdab_{j}^t  +\rho (\xb_{j}^{t+1} -\xb_{0}^t) )\big\|^2
+  \frac{L^2}{2L-1} \nu\big) \notag \\
\overset{\text{$(c)$}}{\leq}& \, (L^2+\rho^2)\|\xb_{j}^t - \xb_{j}^{t+1} \|^2 + \frac{L^2}{4L-2} \nu,
\end{align}where   $(a)$ holds due to the fact that $\xb^{\top}\yb \leq \frac{1}{2L}\|\xb\|^2 +\frac{L}{2} \|\yb\|^2$;   $(b)$ follows because of \eqref{2022-5-5-2200};  $(c)$ holds  because      the fact that $\|\ab\|^2 \leq 2(\|\bb\|^2+\|\cb\|^2)$ when $\|\ab\| \leq\|\bb\|+\|\cb\|$ and the Assumption~\ref{Ass: Assumption1}.

Next,  $E_2$ term can be bounded as follow, 			
\begin{align} \label{Lemma2_2}
&\,\big\| \nabla_{\xb_{0}}\mathcal{L}( \{\xb_{i}^{t}\}, \xb_{0}^{t}, \{\lambdab_{i}^{t}\} ) \big\|^2  \notag\\
=&\,\Big\| \mathbf{x}_{0}^{t} - {\rm prox}_{\rho^{-1} h}\Big( \frac{1}{N} \sum_{i=1}^{N}  \big(\xb_{i}^{t}-\frac{1}{\rho}\lambdab_i^{t} + \xi_i^{t} \big) \Big) \Big\|^2 \notag\\
=&\, \Big\| \mathbf{x}_{0}^{t} -\xb_{0}^{t+1} +\xb_{0}^{t+1} -  {\rm prox}_{\rho^{-1} h}\Big( \frac{1}{N} \sum_{i=1}^{N}  \big(\xb_{i}^{t}-\frac{1}{\rho}\lambdab_i^{t} + \xi_i^{t} \big) \Big) \Big\|^2 \notag\\
\overset{\text{$(a)$}}{\leq}&\, 2\big[\| \mathbf{x}_{0}^{t} -\xb_{0}^{t+1}\|^2 +\Big\|{\rm prox}_{\rho^{-1} h}\Big( \frac{1}{N} \sum_{i=1}^{N}  \big(\xb_{i}^{t+1}-\frac{1}{\rho}\lambdab_i^{t+1} + \xi_i^{t+1})\Big)  - {\rm prox}_{\rho^{-1} h}\Big( \frac{1}{N} \sum_{i=1}^{N}  \big(\xb_{i}^{t}-\frac{1}{\rho}\lambdab_i^{t} + \xi_i^{t} \big) \Big) \Big\|^2 \big] \notag\\
\overset{\text{$(b)$}}{\leq}&\,  2\big[\| \mathbf{x}_{0}^{t} -\xb_{0}^{t+1}\|^2 +\rho \big\|\frac{1}{N} \sum_{i=1}^{N}  \big(\xb_{i}^{t+1}-\frac{1}{\rho}\lambdab_i^{t+1} + \xi_i^{t+1} \big) -\frac{1}{N} \sum_{i=1}^{N}  \big(\xb_{i}^{t}-\frac{1}{\rho}\lambdab_i^{t} + \xi_i^{t} \big) \big\|^2 \big]\notag\\
\overset{\text{$(c)$}}{\leq}&\, 2\| \mathbf{x}_{0}^{t} -\xb_{0}^{t+1}\|^2 +\frac{4N}{N^2} \rho \sum_{i=1}^N \Big( \| \xb_{i}^{t+1} - \xb_{i}^t\|^2+\frac{1}{\rho^2}\|\lambdab_i^t - \lambdab_i^{t+1}\|^2 \Big)
+ 8\rho \| \xi_{i} \|^2 \notag\\
\overset{\text{$(d)$}}{\leq}&\, 2\| \mathbf{x}_{0}^{t} -\xb_{0}^{t+1}\|^2 + 8\rho \| \xi_i \|^2 + \frac{4}{N}\sum_{i=1}^N \Big( (\rho+\frac{4L^2}{\rho})\|\xb_i^{t+1}-\xb_i^t\|^2 +\frac{16}{\rho}\|\varepsilon_i\|^2
+ \frac{8}{\rho}\|e_i\|^2\Big),
\end{align}where  $(a)$ follows because  $\|\ab\|^2 \leq 2(\|\bb\|^2+\|\cb\|^2)$   and the update of $\xb_{0}$ (cf. \eqref{eqn:x_0_update_b}); $(b)$ holds from the non-expansive property  of proximal operater \cite{parikh2014proximal}, i.e.,
\begin{align}
\|\operatorname{prox}_{\alpha_{_U} R} (\xb) - \operatorname{prox}_{\alpha_{_U} R} (\mathbf{y})\| \leq \alpha_{_U}\| \xb - \mathbf{y}\|.
\end{align}
In $(c)$ we use the fact that $\|\ab\|^2 \leq N \sum_{i=1}^{N} \|\bb_i\|^2$;   $(d)$ follows because   the same reasoning as the inequality \eqref{Lemma1_3}.

Finally, we  bound the $E_3$ term in \eqref{eqn:criterion_appendix} as follow,
\begin{align} \label{Lemma2_3}
\| \nabla_{\lambdab_{j}}\mathcal{L}( \{\xb_{i}^{t}\}, \xb_{0}^{t}, \{\lambdab_{i}^{t}\} )  \|^2
=& \|\xb_{0}^t - \xb_j^t\|^2 \notag\\
\overset{\text{$(a)$}}{=}& \frac{1}{\rho^2}\| \lambdab_{j}^{t+1} - \lambdab_{j}^t\|^2  \notag \\
\overset{\text{$(b)$}}{\leq}& \frac{4L^2}{\rho^2} \|\xb_j^{t+1}-\xb_j^t\|^2 +\frac{16}{\rho^2}\|\varepsilon_j\|^2 + \frac{8}{\rho^2}\|e_j\|^2,
\end{align}where $(a)$ follows due to \eqref{eq: sadmm-update_3}; in (b) we use the same reasoning as the inequality \eqref{Lemma1_3}. By combining \eqref{Lemma2_1}, \eqref{Lemma2_2} and \eqref{Lemma2_3}, we obtain
\begin{align}
&\,\mathbb{E}\Big[P(\{\xb_{i}^{t}\}, \xb_{0}^{t}, \{\lambdab_{i}^{t}\})  \Big] \notag \\
\leq&~ 2 \| \mathbf{x}_{0}^{t} -\xb_{0}^{t+1}\|^2 +  \sum_{i=1}^{N} \Big\{\Big[(L^2+\rho^2) +\frac{4}{N}(\rho+\frac{4L^2}{\rho}) +\frac{4L^2}{\rho^2} \Big] \cdot \| \xb_{i}^{t+1} -\xb_i^t\|^2 \notag \\
&  +( 1+\frac{4\rho}{N})\frac{16}{\rho^2} \phi^2 	+ \big(\frac{L^2}{4L-2}+(1+\frac{4\rho}{N})\frac{8}{\rho^2}\big)  \nu \Big\} + 8\rho \sigma^2.
\end{align}
Thus, we complete the proof.
$\hfill\blacksquare$

\bibliographystyle{IEEEtran}
\bibliography{reference}

\end{document}